\documentclass{article}

\usepackage[margin=1in]{geometry}
\usepackage{times}
\usepackage{natbib}
\usepackage{graphicx}
\usepackage{float}
\usepackage{tikz}
\usepackage{algorithm}
\usepackage{algpseudocode}
\usepackage{booktabs}
\usepackage{subcaption}
\usepackage{amsmath}
\usepackage{amssymb}
\usepackage{hyperref}
\usepackage{url}
\usepackage{authblk}
\usepackage{caption}

\title{Visualizing the World Model in an Agent's Value Function}

\author[1]{Shawheen Naderi\thanks{Corresponding author: synaderi@ucdavis.edu}}
\author[1]{Fatemeh Akherati}
\author[1]{Yubei Chen}

\affil[1]{University of California, Davis}

\begin{document}

\maketitle

\begin{abstract}
\noindent Could model-free reinforcement learning algorithms actually be learning a model of the environment? In this work, we apply counterfactual interventions which hold everything about a state fixed, besides a targeted change, to explore if the value function of a Proximal Policy Optimization agent forms a world model of the Crafter environment. Despite only being trained to optimize reward, we discover that the value function learns concepts such as the rules governing when a diamond is collectible, water and food having an impact on health, heuristics regarding inventory, and more. Finally, we discuss the safety implications of our visualizations and some shortcomings of current reinforcement learning agents that will prevent us from building truly advanced machine intelligence.
\end{abstract}

\section{Introduction}

How do humans -- and other animals -- determine which actions to take in a given scenario? An answer may lie in the ability to observe a state of the world and then predict its future consequences -- including rewards or punishments -- without needing the future to actually unfold ~\citep{ilya_were_2025}. One can then take the action that leads to the best expected state. For instance, a person navigating in a new city may attempt a route, recognize that they are headed into a dead end, and change directions before finishing their initial route. After encountering similar situations, they learn to recognize these patterns and drive more efficiently.

The above process is simplified in machine agents through the \textit{value function}, which estimates the discounted expected sum of future rewards obtained by following actions selected by the \textit{policy} from a given state ~\citep{sutton_reinforcement_2014}. The value function's predictions are then used to update the policy to select more rewarding actions. Mathematically, this is written as
$v^\pi(s)=\mathbb{E}_\pi\!\left[\sum_{k=0}^{\infty}\gamma^k R_{t+k+1}\,\middle|\,S_t=s\right].$ In practice, the value function is predicted using a neural network called the \textit{critic}, and often the critic is referred to as the value function since it aims to model it ~\citep{konda_actor-critic_2000}. We argue that in order for the value function to be effective, the critic develops an implicit \textit{world model} ~\citep{sutton_dyna_1991, ha_world_2018, hafner_dream_2020}.  A world model allows an agent to take in past inputs and predict the future which requires internalizing the dynamics of an environment. Implicit is used to describe that the agent is not explicitly being optimized for this task but instead the world model emerges as a byproduct of estimating return. \textbf{The world model residing in an agent's value function can be investigated through \textit{counterfactual analysis} by slightly altering an observed state and recording changes in the value function's predictions.} This is critical, as understanding how an agent perceives the environment is essential to guarantee it behaves safely, is aligned with human interests, and also improve upon its weaknesses.

Our experiments will analyze the critic from a Proximal Policy Optimization (PPO) algorithm trained in the Crafter environment -- a simplified 2D version of Minecraft ~\citep{hafner_benchmarking_2022,schulman_proximal_2017}. Crafter is an ideal testbed for our method because the hierarchical task structure of this environment requires the agent to develop several skills and then generalize to new maps in each episode. Additionally, its simplicity allows for quick iterations and long-horizon training. 

We observe that, after many training steps, the value function does indeed form a world model. The main aspects of the world that a veteran player should have learned are: \textbf{1) that the world contains finite reward, 2) the requirements to collect specific achievements, and 3) what causes increases or decreases in health}. We confirm that the agent has learned all of these by employing counterfactual analysis. A heuristic we find is that the inventory component of an agent's observation -- which contains tools like swords or raw materials like wood -- is employed as a measure of progress in an episode and significantly influences value predictions indicating the value function has learned about the environment's finite reward. Impressively, the value function is able to learn when achievements -- such as collecting the diamond -- are possible by disentangling their mere presence from their actual attainability which typically requires tools like pickaxes.  Finally, we determine that the value function has understood what leads to health increases or decreases by assessing resources like water and food depending on its needs.

\section{World Structure of Crafter}

\begin{figure}[H]
    \centering
    \hspace*{-0.03\textwidth}%
    \begin{subfigure}[t]{0.64\textwidth}
        \centering
        \vspace{0pt}
        \includegraphics[
            width=\linewidth,
            height=5.2cm
        ]{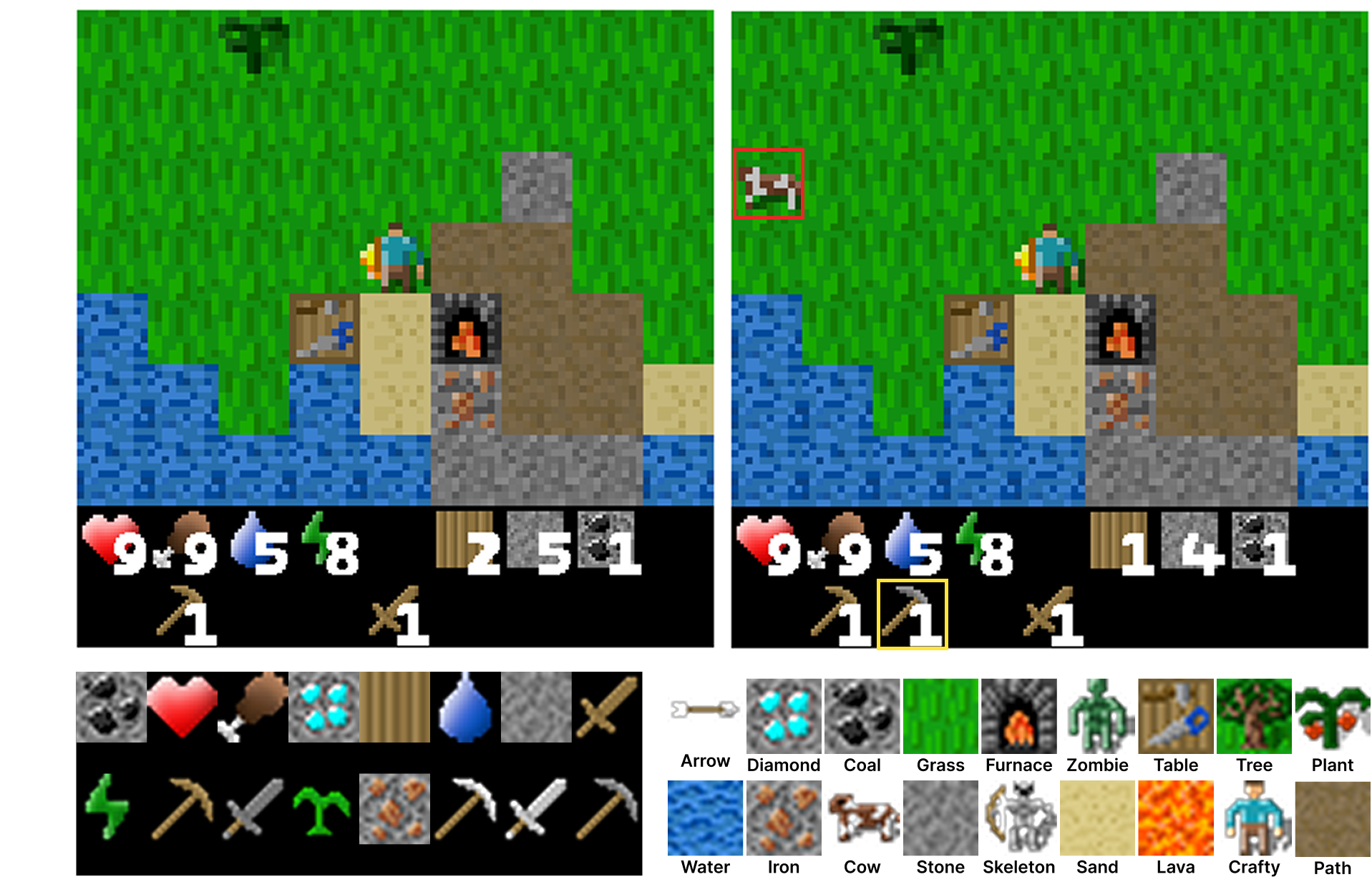}
        \caption{Crafting a stone pickaxe.}
        \label{fig:crafter_environment}
    \end{subfigure}%
    \hspace{0.02\textwidth}%
    \begin{subfigure}[t]{0.34\textwidth}
        \centering
        \vspace{0pt}
        \includegraphics[
            width=\linewidth,
            height=5.2cm
        ]{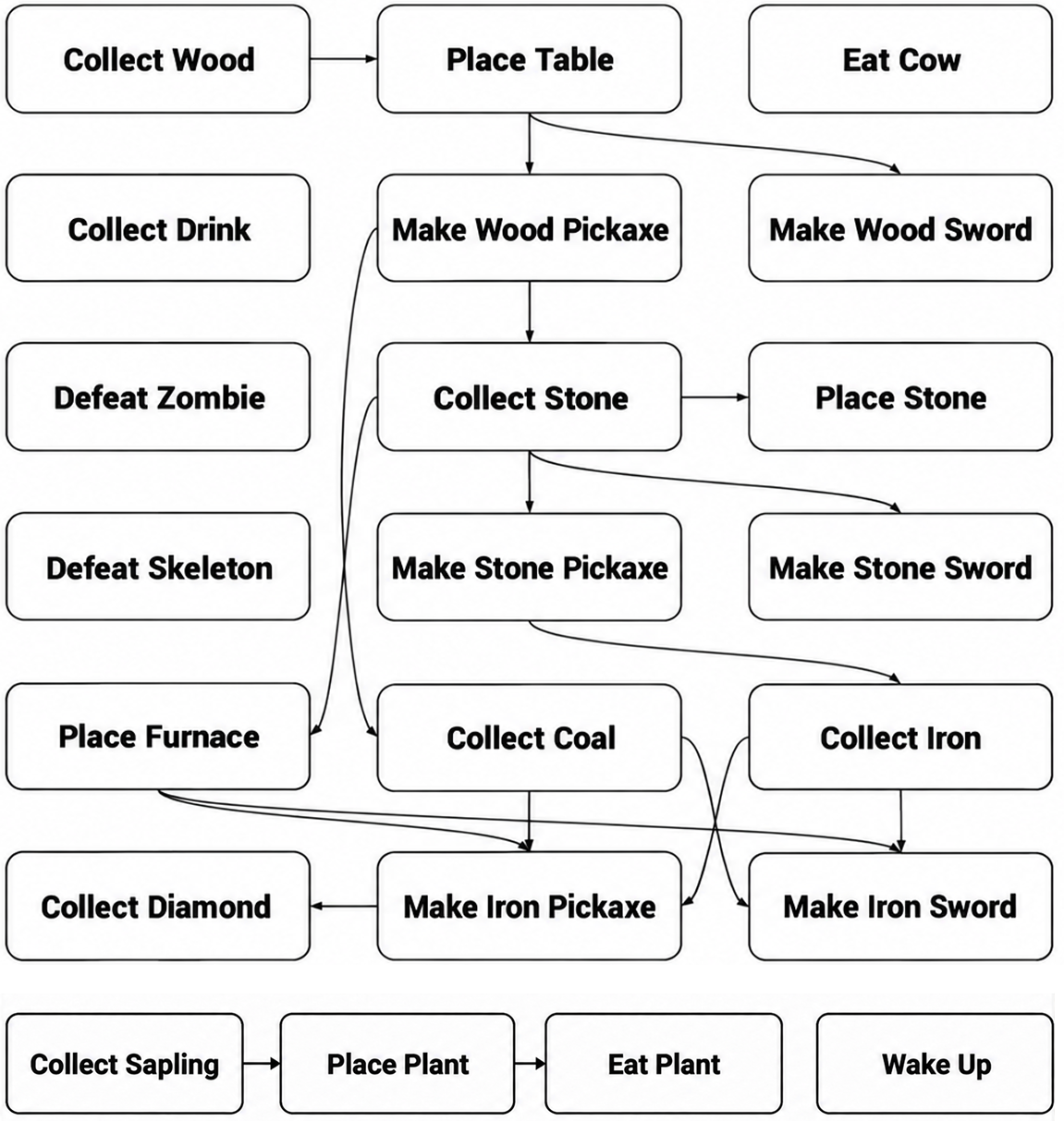}
        \caption{Achievement hierarchy.}
        \label{fig:achievement_map}
    \end{subfigure}

    \caption{
    In panel (a), the pair shows the agent before and after crafting a stone pickaxe which is highlighted by the yellow box, and a cow also appears highlighted by the red box. Crafting the pickaxe requires the agent to be near a table and possess the necessary wood and stone resources; afterward, the stone pickaxe appears in the inventory, enabling the agent to collect iron. The external environment is composed of 18 possible textures, while the internal environment contains 16 possible textures including inventory items and health-related components. The agent’s view of the environment is only partially observable. Panel (b) shows the achievement graph, where arrows indicate prerequisite relationships between achievements.
    }
    \label{fig:crafter_overview}
\end{figure}

Crafter was developed as a research environment for fast iteration in reinforcement learning. An agent plays episodes (we sometimes refer to these as ``lives") where it needs to complete tasks such as collecting resources, fighting creatures, and crafting tools until eventually mining a diamond. To avoid redundancy, we name the agent Crafty. When Crafty collects a resource -- or builds a tool -- those items are placed in its inventory. The agent has a health component that depends on food, water, and energy. When Crafty runs out of one of these its health slowly degenerates or, when it has all three of these, its health gradually rejuvenates. Combat with creatures like zombies or skeletons can also degrade health. When the agent exhausts its health -- a maximum of 9 units -- the episode terminates and a new episode begins. Each episode is procedurally generated so the agent is required to generalize and not simply memorize the same map.

\subsection{Hierarchical Achievements}
The game consists of 22 hierarchical achievements. The first time an agent completes an achievement, it receives a reward of 1, allowing it to accumulate a total reward of at most 22 in a single episode. Additionally, changes in health are rewarded by ±0.1, with positive rewards for health increases and negative rewards for health decreases. Some achievements are purely internal, like waking up, but most require the agent to interact with the external environment to find and then gather the resources needed to obtain them. There is also a strict prerequisite for each achievement, meaning that it cannot collect ``achievement B" until it has first collected ``achievement A." For example, the agent is unable to collect stone until it has the wood pickaxe in its inventory. 

\subsection{Actions}
The agent may take 17 different actions in the environment. Some actions correspond to multiple outcomes such as ``Do", which can kill other creatures, collect objects for its inventory, etc. but other actions like ``Make Iron Sword" are specific to one task. If an agent attempts an action before it has the prerequisite achievement, then nothing will happen (see Appendix~\ref{app:Actions} for a table of actions). 

\section{Experimental setup}
We train a standard PPO agent for 100 billion environment steps in the Craftax-Classic environment, a Jax implementation of Crafter optimized for speed ~\citep{matthews_craftax_2024}. Throughout training, checkpoints are saved and the checkpoint used in experiments is saved at 10 billion steps because this is a conservative point where the marginal improvement begins to diminish as steps are scaled. A symbolic, rather than pixel, observation is used to further accelerate training, after which it is rendered to pixels for visualization purposes. No problems should arise translating between pixel textures and symbolic states since the pixel textures are already discrete entities re-used throughout the environment like building blocks. Once fully trained, Crafty is able to obtain almost every achievement each episode but surprisingly still struggles with the two most complex achievements since they are last on their respective achievement graph -- eat plant and collect diamond (see Appendix~\ref{app:Expiremental_setup} for details on hyper-parameters, the architecture, and Table~\ref{tab:achievement_success_over_training} which shows success rates over training time). 

With the checkpoint, we collected 10,000 new episodes to construct an evaluation dataset for subsequent experiments. The dataset contained 9,436,937 transitions, with a median episode length of 632 steps; only five episodes reached a 10,000-step cap. Crafty collected a diamond in $35.0\%$ of episodes. Overall, $69.0\%$ of episodes ended in lava, rising to $78.0\%$ among episodes where diamond was collected, compared with $64.2\%$ among episodes without a diamond.

\section{Method}
\subsection{Counterfactual Interventions in Crafter}
Counterfactual analysis is an approach that considers how an outcome would differ if specific components were modified while keeping everything else the same ~\citep{pearl_book_2020}. The goal of counterfactual analysis is to separate causal effects from correlations. Counterfactuals have been applied in reinforcement learning but have focused primarily on the policy rather than the value function ~\citep{amitai_explaining_2023, gajcin_redefining_2024, olson_counterfactual_2021, mesnard_counterfactual_2021}. Crafter is particularly suited for counterfactual analysis because its symbolic state structure allows us to precisely alter aspects of the environment and document changes in the value function's predictions using a model with frozen parameters. We can modify the internal and external states, or both simultaneously, to test interactions. This enables us to determine if the value function is simply learning correlations with reward or if it has also discovered structured representations reflecting aspects of the environment's transition dynamics. In the diamond example, if the critic had only learned correlations, the presence of the diamond would increase the value prediction because it is associated with a positive reward. However, if the value function learned deeper rules about the environment, then the mere addition of a diamond would not significantly increase value unless it is actually exploitable, i.e., the diamond is present and the agent has the iron pickaxe.

\subsection{Expected Orderings}
 Counterfactual analysis can be further systematized by introducing a set of orderings on these counterfactuals that act as a hypothesis on what the value function has learned and then a success rate on the percent of orderings correct can be employed to review that the value function indeed has converged to that ordering. 
A sequence of different counterfactuals can be performed; then, domain expertise can rank which state should have the highest predicted value. For instance, a state with a diamond that can be collected for the first time should usually be ranked higher than an alternative state without a diamond present if the value function has truly internalized the constraints of the environment. We call this the \textit{expected ordering} of a state's value. 

\subsection{Algorithm}
To use our expected ordering, we select various states with a desired property, generate counterfactuals on these states, and calculate the success rate as a proportion of the orderings that the value function matched for the expected ordering. If the agent has a high success rate, then its value function has likely developed a strong internal model of the concept. 

\begin{algorithm}[htbp]
\caption{Evaluating Expected Ordering Success Rate}
\label{alg:platonic_ordering}
\begin{algorithmic}[1]
\Require Value function $\hat V$, set of states $\mathcal{S}$, counterfactual operator $\mathcal{C}$, expert ordering rule $\prec^*$
\Ensure Success rate of the expected ordering

\State Initialize $N \gets 0$ and $K \gets 0$

\For{each state $s \in \mathcal{S}$}
    \State Generate counterfactual states: $\{s_1, s_2, \dots, s_m\} \gets \mathcal{C}(s)$
    \State Use domain knowledge to define the expected ordering: $\
        s_1 \prec^* s_2 \prec^* \cdots \prec^* s_m$
    \State Evaluate the value function on each counterfactual:$\
        \hat V(s_1), \hat V(s_2), \dots, \hat V(s_m)$
    \If{$\hat V(s_1) < \hat V(s_2) < \cdots < \hat V(s_m)$}
        \State $K \gets K + 1$
    \EndIf
    \State $N \gets N + 1$
\EndFor

\State \Return $\text{SuccessRate} = K/N$
\end{algorithmic}
\end{algorithm}

\section{Discoveries}

\subsection{The Value Function Has Learned About the World's Finite Reward and Uses Inventory as a Heuristic to Measure Progress}

\begin{figure}[htbp]
    \centering
    \includegraphics[width=1.0\linewidth]{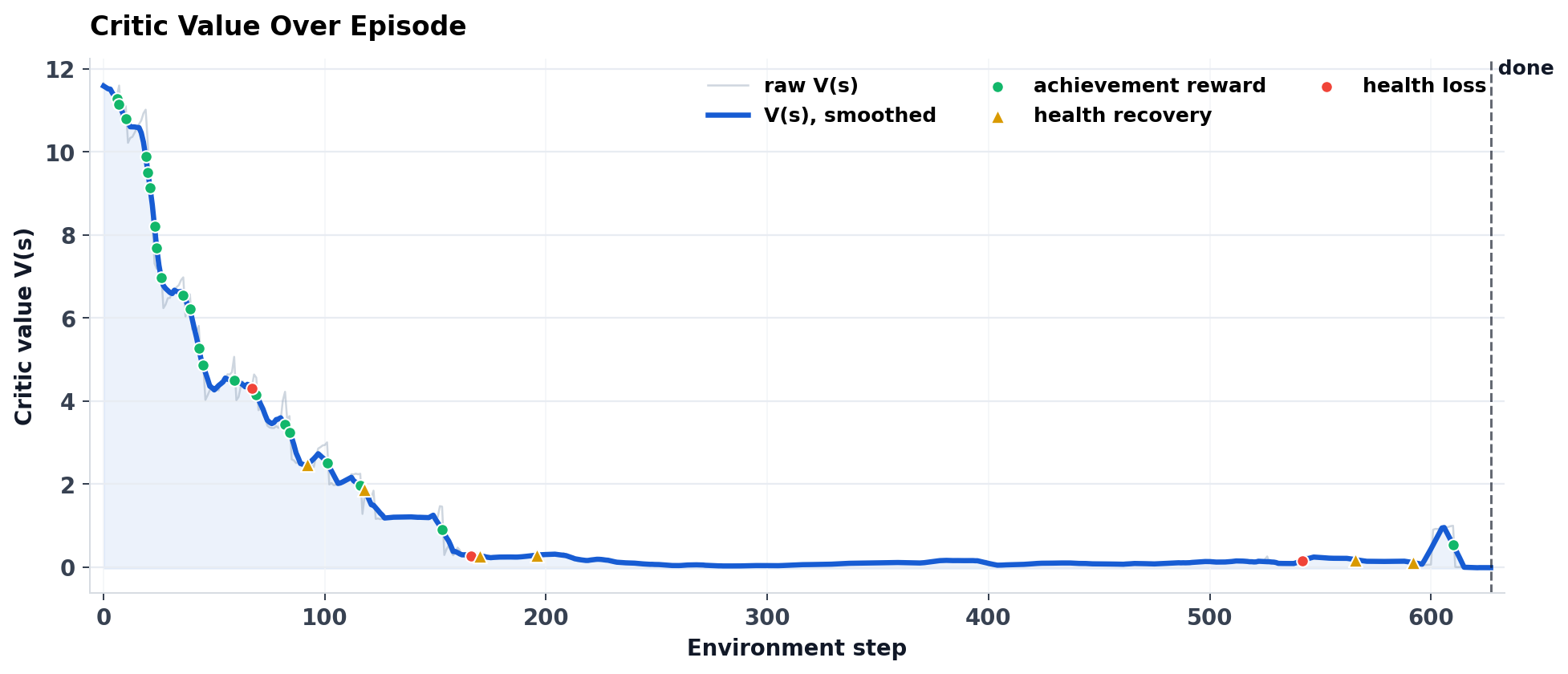}
    \caption{The  curve shows the raw predictions, the blue curve shows the smoothed value estimate, and the markers indicate achievement, health, and negative rewards. The predicted value declines rapidly as achievements are collected and remains near zero for most of the later episode, with a brief increase near the end when a diamond is collected.}
\end{figure}

{\noindent \bf The Value Function Collapses to Near Zero as the Episode Progresses.} We can now apply counterfactual analysis to uncover Crafty's learned heuristics and truly understand what is happening inside the value function. First, we can simply plot the value function's predictions at each step across an episode that was played using our final frozen network. Surprisingly, we discover that Crafty's value function makes very high predictions for initial states but rapidly decreases to have flat near zero values as the episode develops. Why is this? Perhaps the agent has recognized that the world has finite rewards and the further it advances the less rewards there are remaining. However, there is no memory mechanism for the agent, so how is it tracking progress? A plot of some different states throughout the episode may help reveal a pattern. 

\begin{figure}[H]
    \centering
    \includegraphics[width=1.00\linewidth]{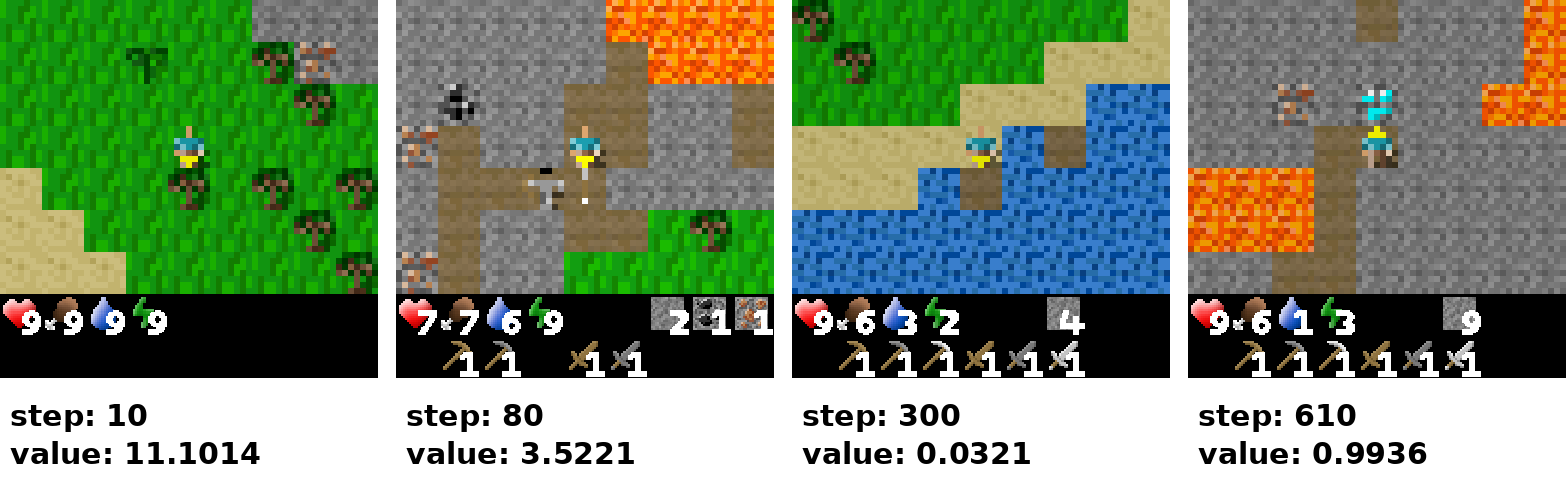}
    \caption{Starting with the earliest steps on the left, we see that Crafty has an empty inventory and has yet to collect any resources. As Crafty continues through the episode, it starts expanding its inventory which allows it to enter cave areas, but also expends the finite rewards in the environment. By step 300, the agent has collected almost all the achievements, excluding diamond, and thus has a very low predicted value until it finds the diamond at step 610.
}
    \label{fig:placeholder}
\end{figure}

{\noindent \bf Hypothesis on Value Function Behavior.} The small bits of Crafty's life we observed hint at a likely possibility: the value function may be using inventory as a tool to account for achievements already collected. When the inventory adds an item, the value function will make a lower prediction signaling it has learned the finite reward of the environment. To test our hypothesis, we will use counterfactual analysis to apply an intervention by solely increasing inventory in a manner that corresponds to the achievement graph and observe if the value function's predictions decrease. 

\begin{figure}[H]
    \centering
    \includegraphics[width=1.00\linewidth]{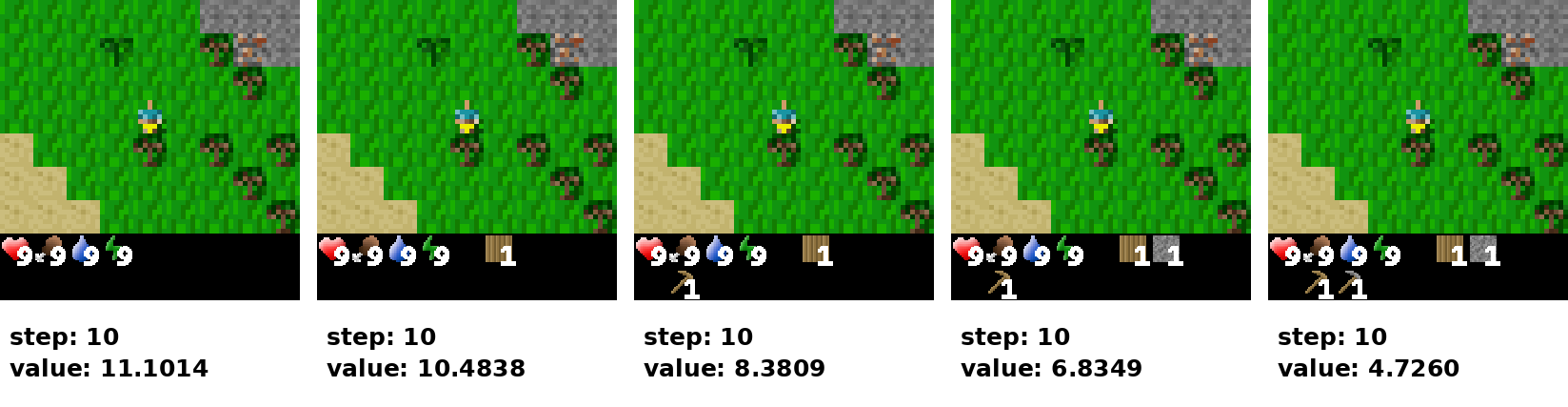}
    \caption{Starting from an unaltered state with an empty inventory, we sequentially add wood, a wood pickaxe, stone, and a stone pickaxe while keeping the rest of the state fixed. The critic's predicted value generally decreases as the inventory progresses. 
}
    \label{fig:placeholder}
\end{figure}

{\noindent \bf Success Rates of Expected Orderings  confirm this trend.}
We sample early- and late-game observations and swap their inventories. If inventory is the dominant signal of game stage, we expect: \textbf{V(late stage observation, late stage inventory) to be less than V(late stage observation, early stage inventory) and V(early stage observation, late stage inventory) to be less than V(early stage observation, early stage inventory)}. Success is defined as the percentage of episode pairs satisfying both inequalities. Late-game inventory lowered the critic's predicted value in both early- and late-game observations for $\mathbf{89.1\%}$ of episode pairs. These results provide further evidence that the critic uses inventory as a signal of game stage and has learned the finite reward of the environment. Full experimental details are provided in Appendix~\ref{app:Experiments}.

{\noindent \bf Depression causes Suicide.} Once Crafty collects the diamond, few rewards remain, and its value function drops to approximately zero. Although health regeneration can provide additional rewards, continued interaction also risks negative rewards from health loss. For further intuition, consider the scenario when the agent has full health and has collected all achievements. The only possibility for future rewards is to regenerate health which would first require losing health.  If a health point is lost at step \(D\) and restored at step \(U>D\), the discounted reward is \(-0.1\gamma^{D-1}+0.1\gamma^{U-1}<0\) for \(\gamma=0.99\). Thus, termination of the episode can be preferable when future rewards do not offset near-term negative rewards. States with negative predicted value can be considered ``depressed states."

Indeed, $69.0\%$ of the episodes in the evaluation dataset were terminated by the agent entering lava, which ends the episode without any negative reward, compared to dying from health degeneration which has a negative reward. Immediately before lava entry, $71.84\%$ of states had negative predicted values, $99.2\%$ had food, water, or energy at two or below, and $57.2\%$ had full health. This suggests that Crafty recognized that health degeneration was near and chose to take the coward's way out.

{\noindent \bf Countering Depression.}
The simplest approach to prevent an agent from repeatably terminating the episode would be to impose a penalty upon it for doing so. When we add a -2 penalty for death, the agent no longer exhibits such actions and thus significantly increases its performance on the two hardest tasks: collect diamond and eat plant.

\begin{figure}[htbp]
    \centering
    \begin{subfigure}[t]{0.37\linewidth}
        \centering
        \includegraphics[width=\linewidth]{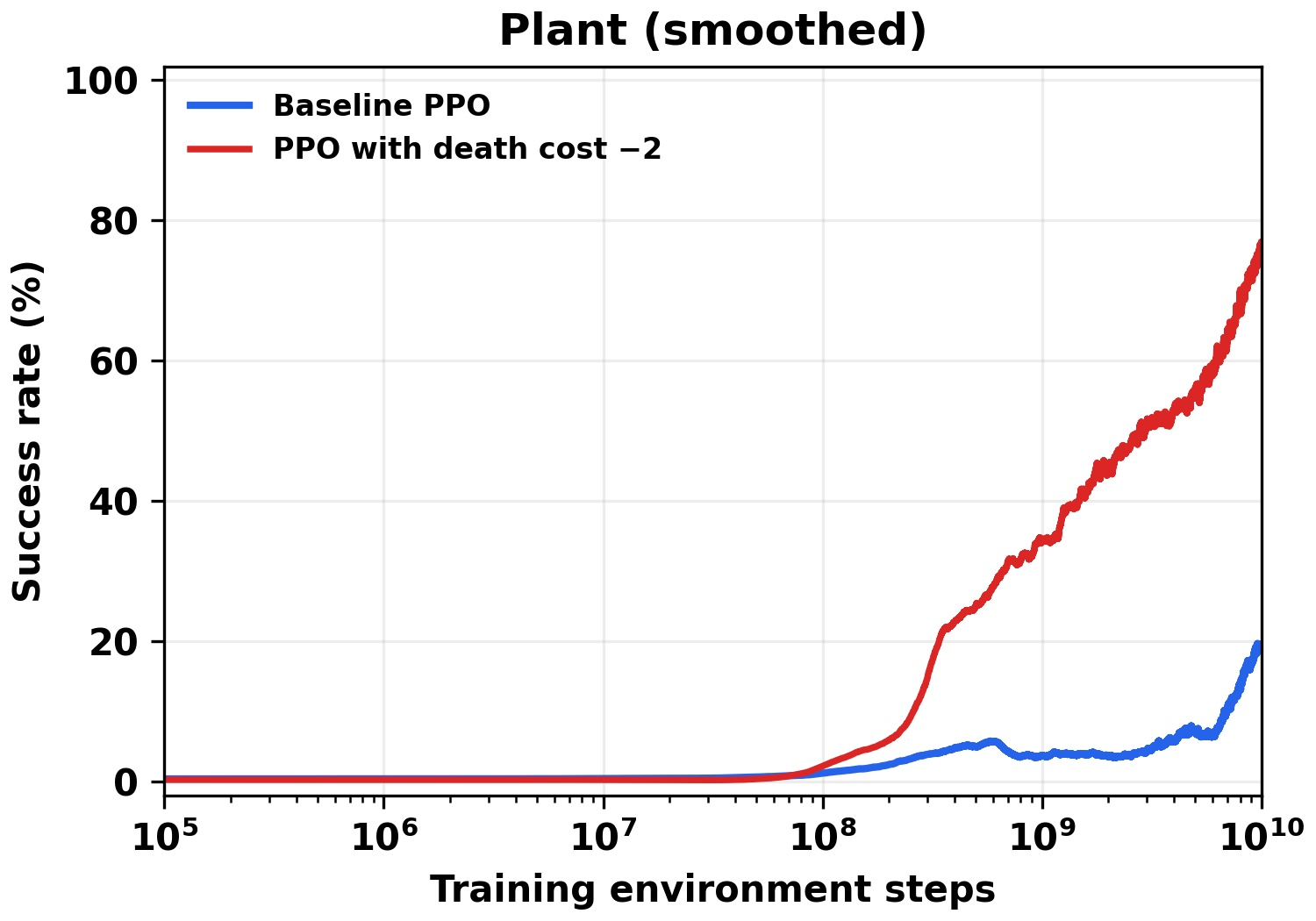}
        \label{fig:plant_penalty}
    \end{subfigure}
    \hspace{0.03\linewidth}
    \begin{subfigure}[t]{0.37\linewidth}
        \centering
        \includegraphics[width=\linewidth]{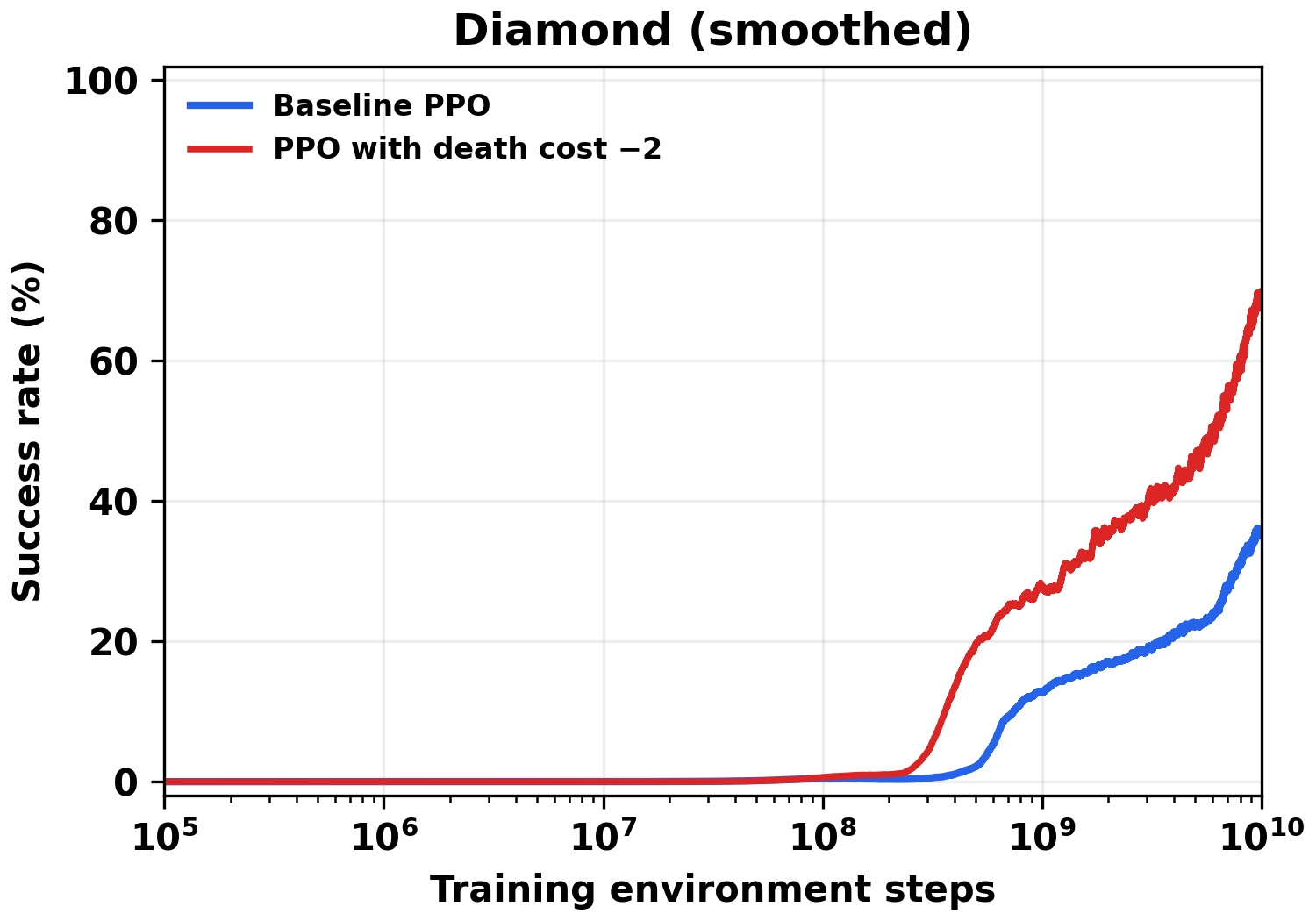}
        \label{fig:diamond_penalty}
    \end{subfigure}

    \caption{Achievement success rates with and without the penalty.}
    \label{fig:achievement_penalties}
\end{figure}

\subsection{The value function has learned the environment's rules governing achievements such as diamond, iron, and plant collectability}
\begin{figure}[htbp]
    \centering

    \begin{subfigure}[t]{0.32\textwidth}
        \centering
        \includegraphics[width=\linewidth]{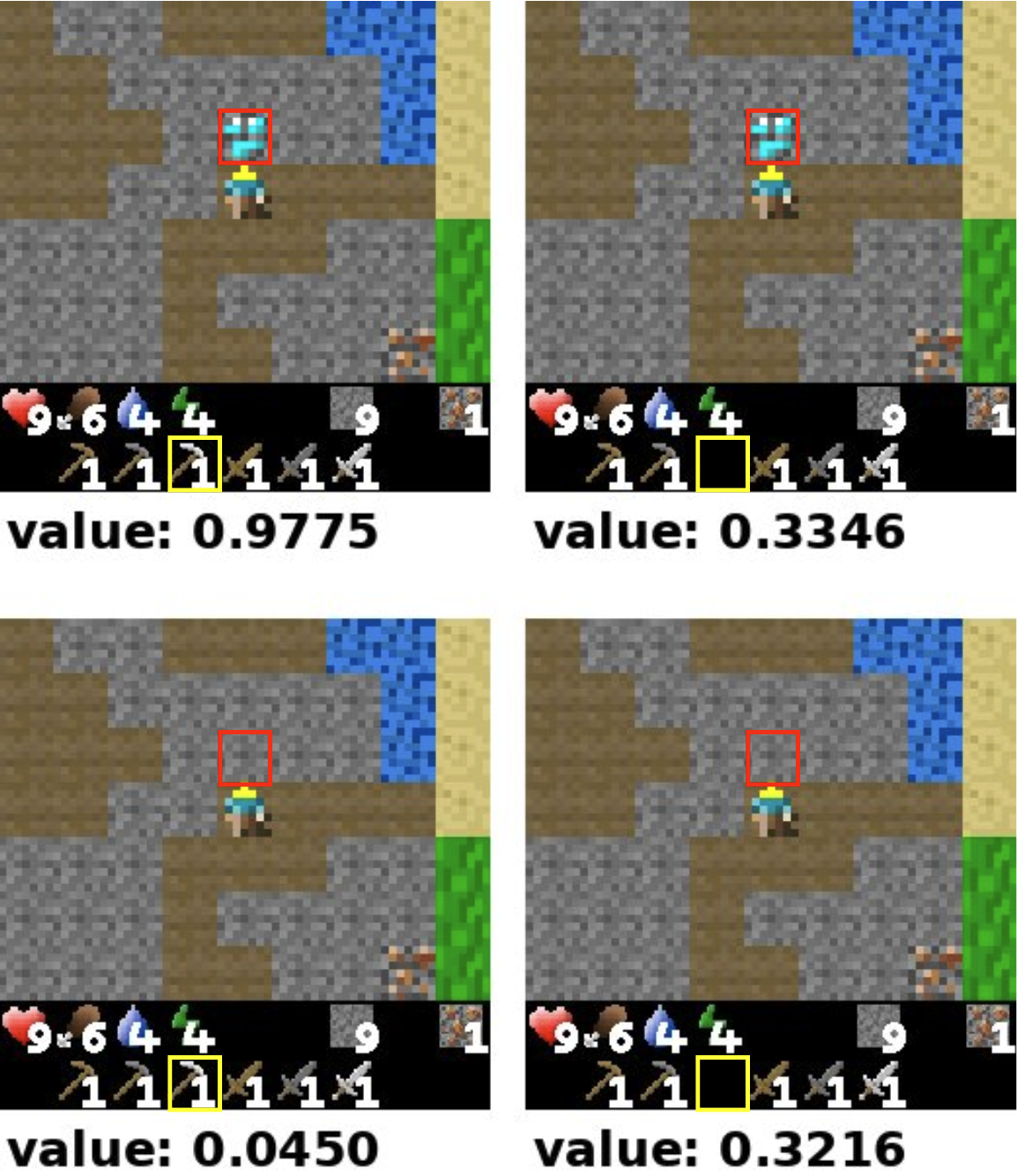}
        \caption{Diamond counterfactual.}
        \label{fig:diamond_counterfactual}
    \end{subfigure}
    \hfill
    \begin{subfigure}[t]{0.32\textwidth}
        \centering
        \includegraphics[width=\linewidth]{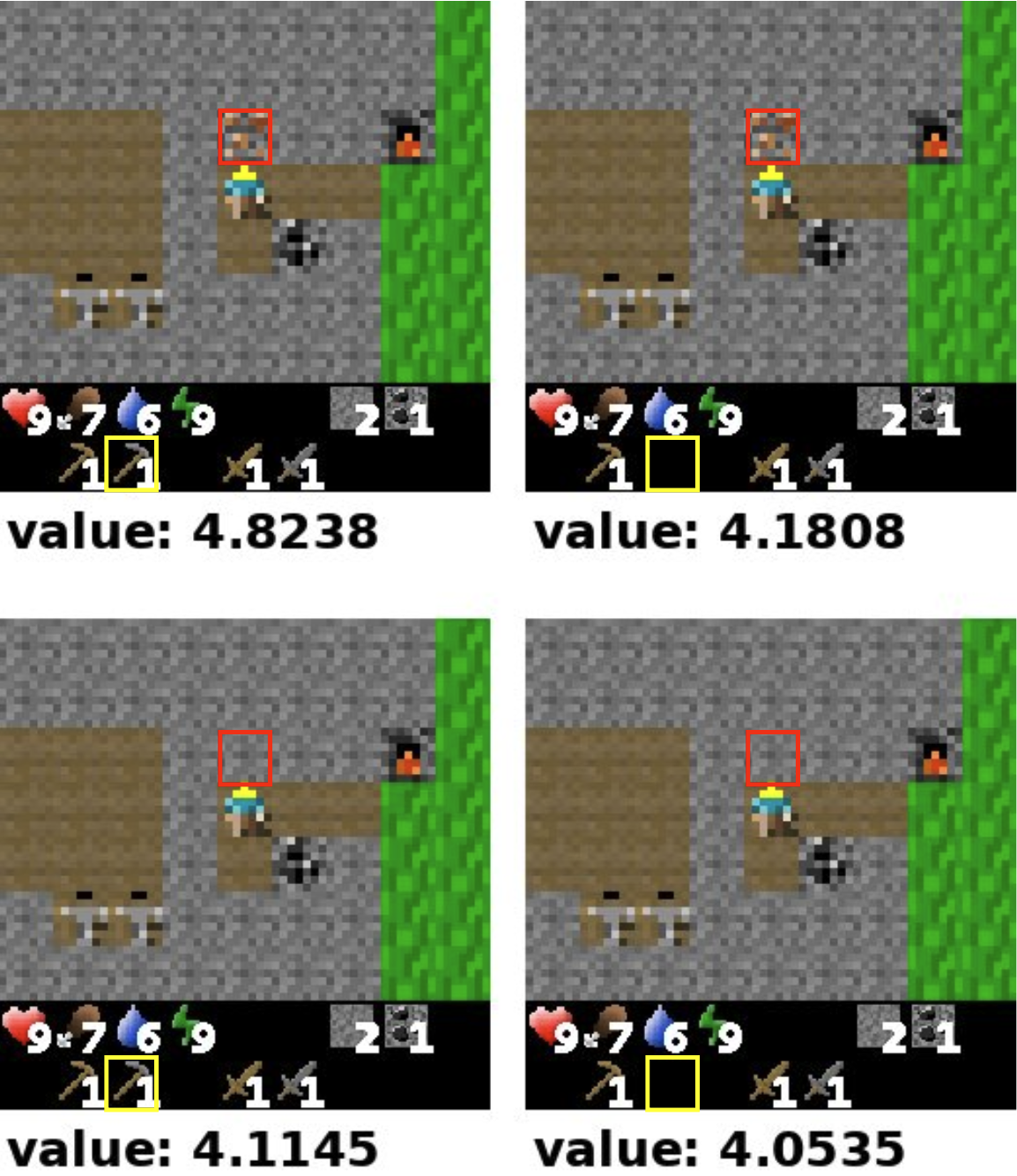}
        \caption{Iron counterfactual.}
        \label{fig:iron}
    \end{subfigure}
    \hfill
    \begin{subfigure}[t]{0.32\textwidth}
        \centering
        \includegraphics[width=\linewidth]{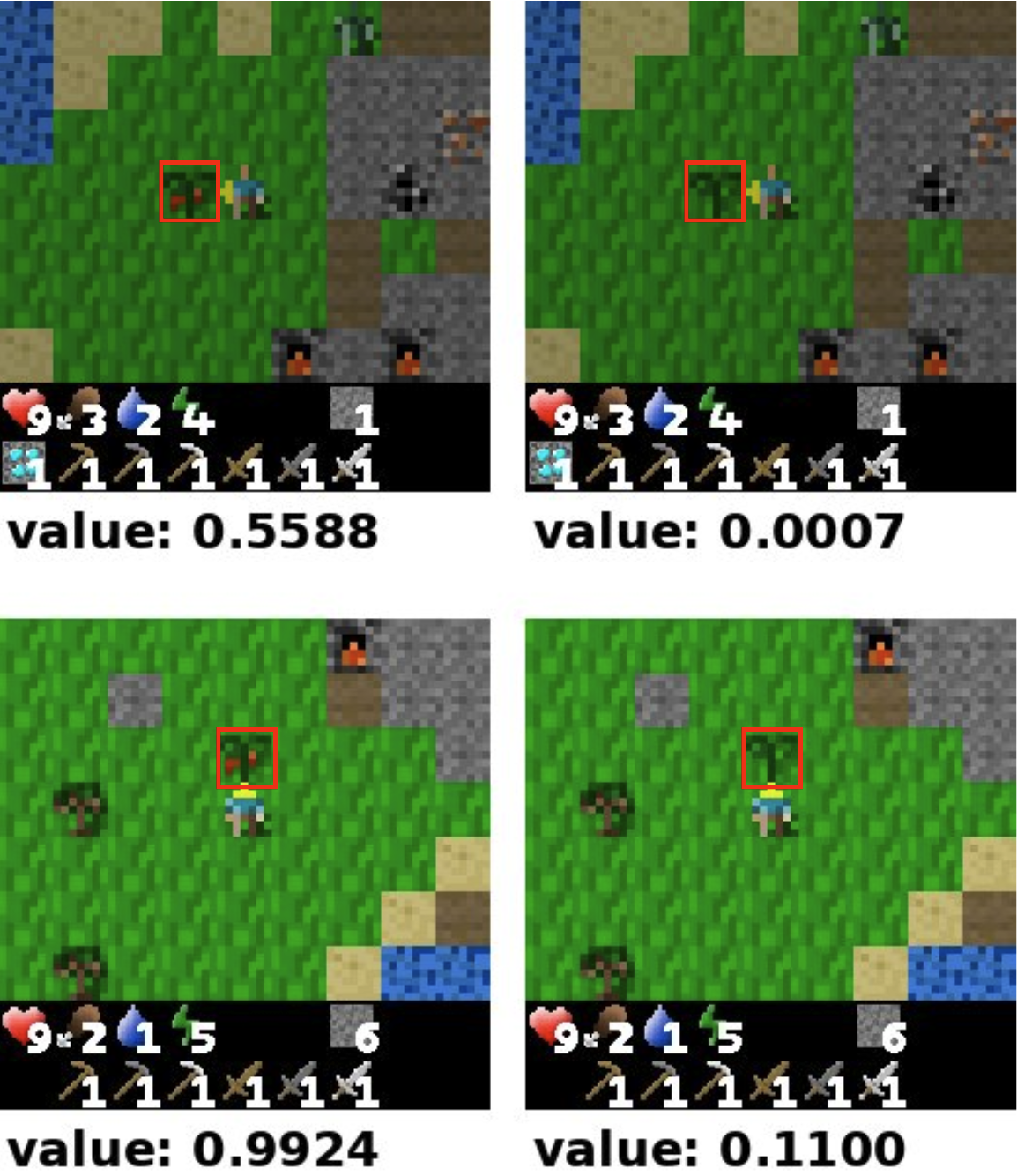}
        \caption{Plants counterfactual.}
        \label{fig:plant}
    \end{subfigure}

    \caption{Three different counterfactual cases for diamond, iron, and plant. Changes are denoted with red and yellow bounding boxes.}
    \label{fig:diamond_examples}
\end{figure}

Our method can also detect that the model has learned the rules regarding to the obtainability of achievements such as diamond, iron, and plant. The diamond can only be collected when Crafty has an iron pickaxe in its inventory, thus, we posit that the value of the state with the diamond plus iron pickaxe present will be higher than all other states and even offset the discount incurred by having an extra item in inventory. To address the immediate confounder that the agent will simply craft an iron pickaxe when a diamond is present but it does not have the iron pickaxe which would result in doubling the reward, we filter the states where its inventory has the resources to do this. Our expected ordering is: \textbf{V(diamond, iron pickaxe) is greater than the maximum of V(diamond, no iron pickaxe), V(no diamond, no iron pickaxe), and V(no diamond, iron pickaxe)} and the success rate is: $\mathbf{95.61\%}$ which provides strong evidence that the value function has learned the concept of diamond collectability and is not merely correlating the presence of an unexploitable diamond with reward. Next, we can perform a very similar experiment to verify the model has learned iron collectability, which requires the agent to have a stone pickaxe, by applying an expected ordering of: \textbf{V(iron, stone pickaxe) is greater than the maximum of V(iron, no stone pickaxe), V(no iron, no stone pickaxe), and V(no iron, stone pickaxe)}. The success rate is  $\mathbf{87.67\%}$  which indicates that the value function has also learned that iron can only be collected with a stone pickaxe. Finally, we test whether the model has learned when it can eat the plant which necessitates the plant to have ripened and is indicated by fruit appearing on it. Thus, we test the predicted value of a ripened plant versus the counterfactual case of an unripe plant with an expected ordering of: \textbf{V(ripe plant) is greater than V(unripe plant)}. The success rate is $\mathbf{99.31\%}$, therefore, we conclude that the model has learned the requirement to eat the plant. Full experimental details are provided in Appendix~\ref{app:Experiments}.

{\noindent \bf Development over training.}
Our approach can obtain an estimate of when a concept has formed during the course of training. To do this, we simply save checkpoints across training steps and evaluate the success rates of the expected orderings at each checkpoint. Early checkpoints where the achievement has never been collected can be noisy, but when Crafty becomes successful at collecting the achievement its expected ordering success rate rises. 

\begin{figure}[H]
    \centering

    \begin{subfigure}[t]{0.32\textwidth}
        \centering
        \includegraphics[width=\linewidth]{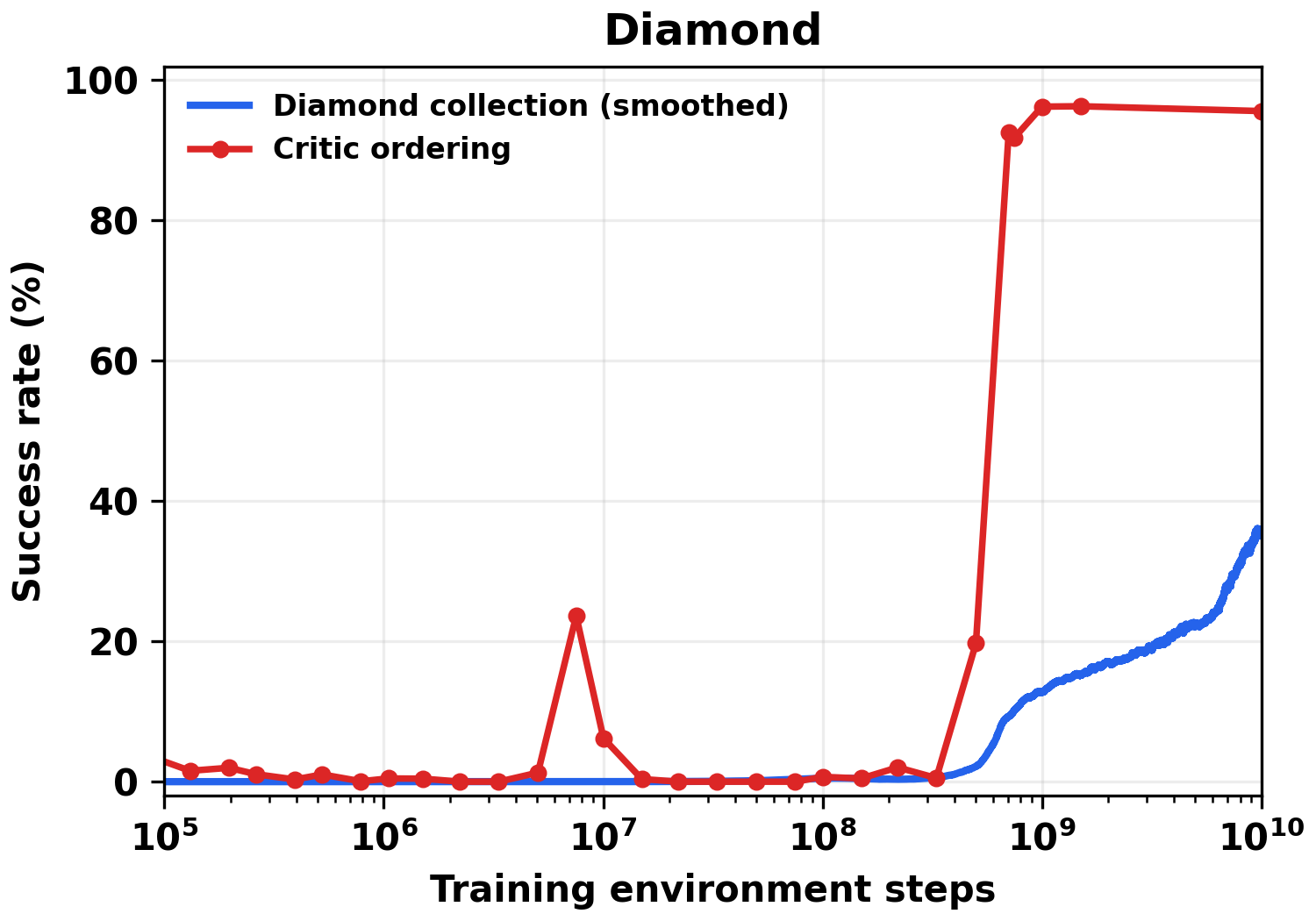}
        \caption{Diamond counterfactual.}
        \label{fig:diamond_counterfactual}
    \end{subfigure}
    \hfill
    \begin{subfigure}[t]{0.32\textwidth}
        \centering
        \includegraphics[width=\linewidth]{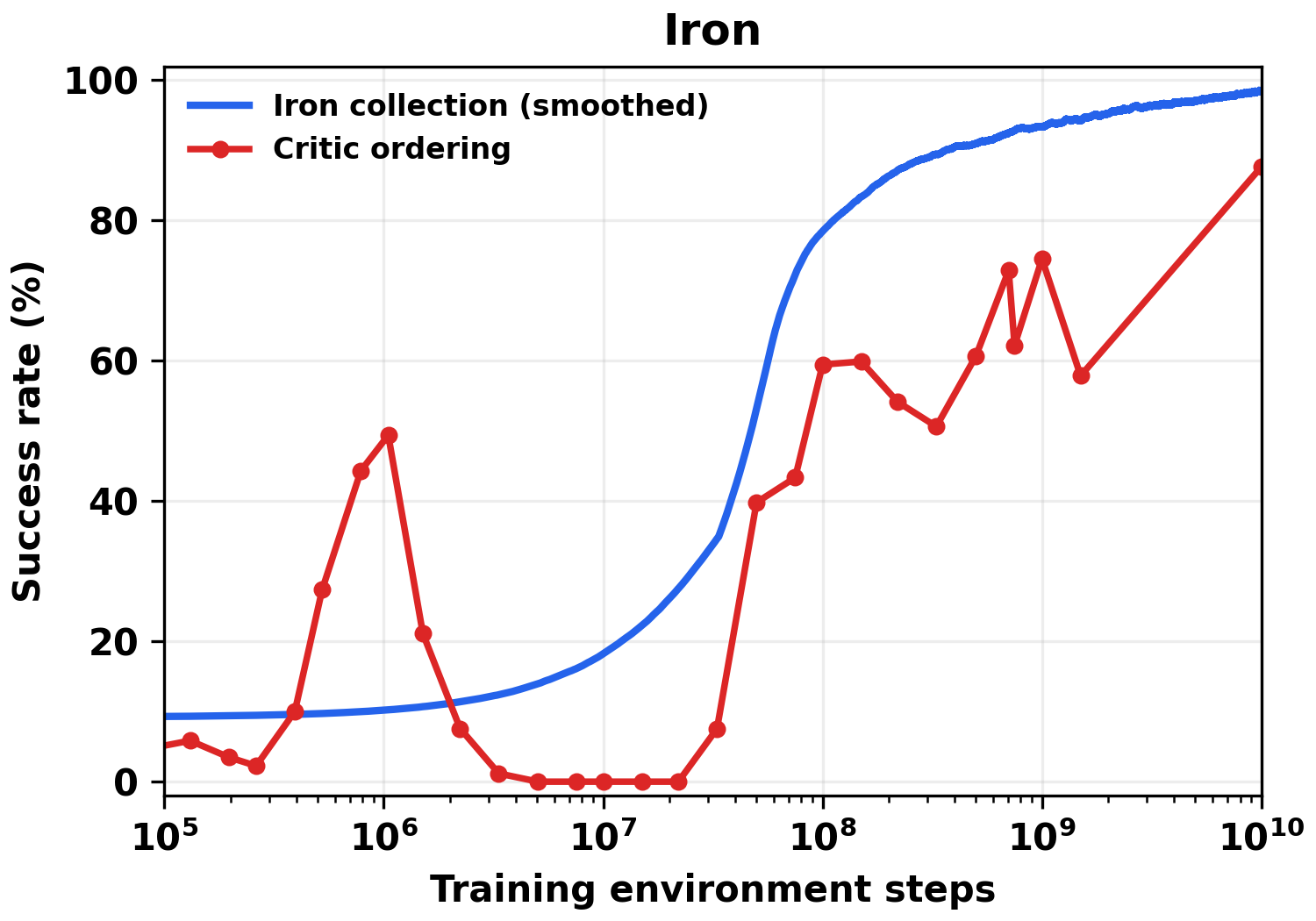}
        \caption{Iron counterfactual.}
        \label{fig:iron}
    \end{subfigure}
    \hfill
    \begin{subfigure}[t]{0.32\textwidth}
        \centering
        \includegraphics[width=\linewidth]{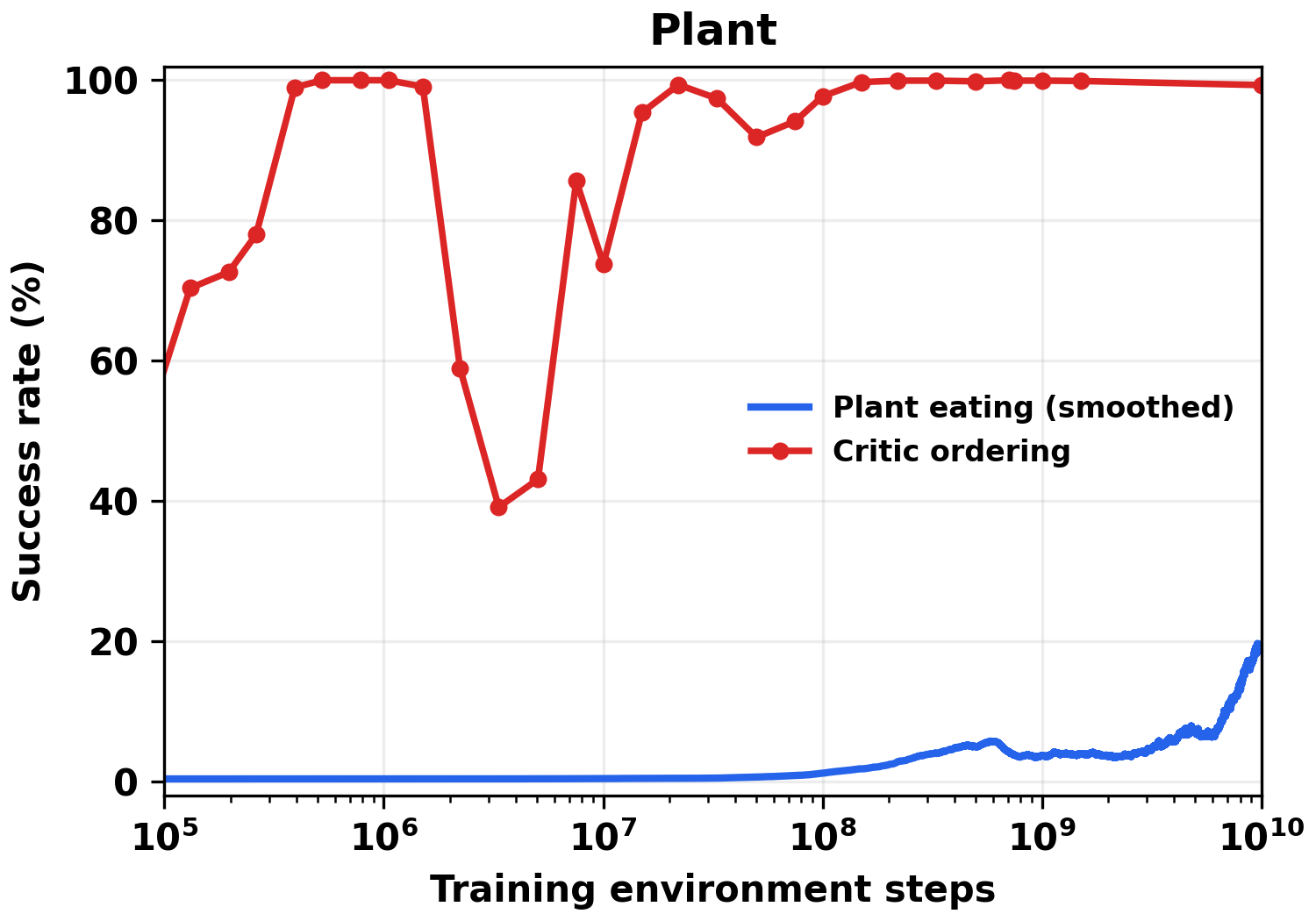}
        \caption{Plant counterfactual.}
        \label{fig:plant}
    \end{subfigure}

    \caption{The blue curve shows the smoothed success rate at which the agent collects the achievement during training and is defined by the percentage of episodes in which the agent successfully collected the achievement at least once. The red curve shows the fraction of fixed counterfactual states for which the critic satisfies the expected ordering.}
    \label{fig:diamond_examples}
\end{figure}

\subsection{The Value Function Has Learned About Key Drivers of Health}
 \begin{figure}[htbp]
     \centering
     \includegraphics[width=0.85\linewidth]{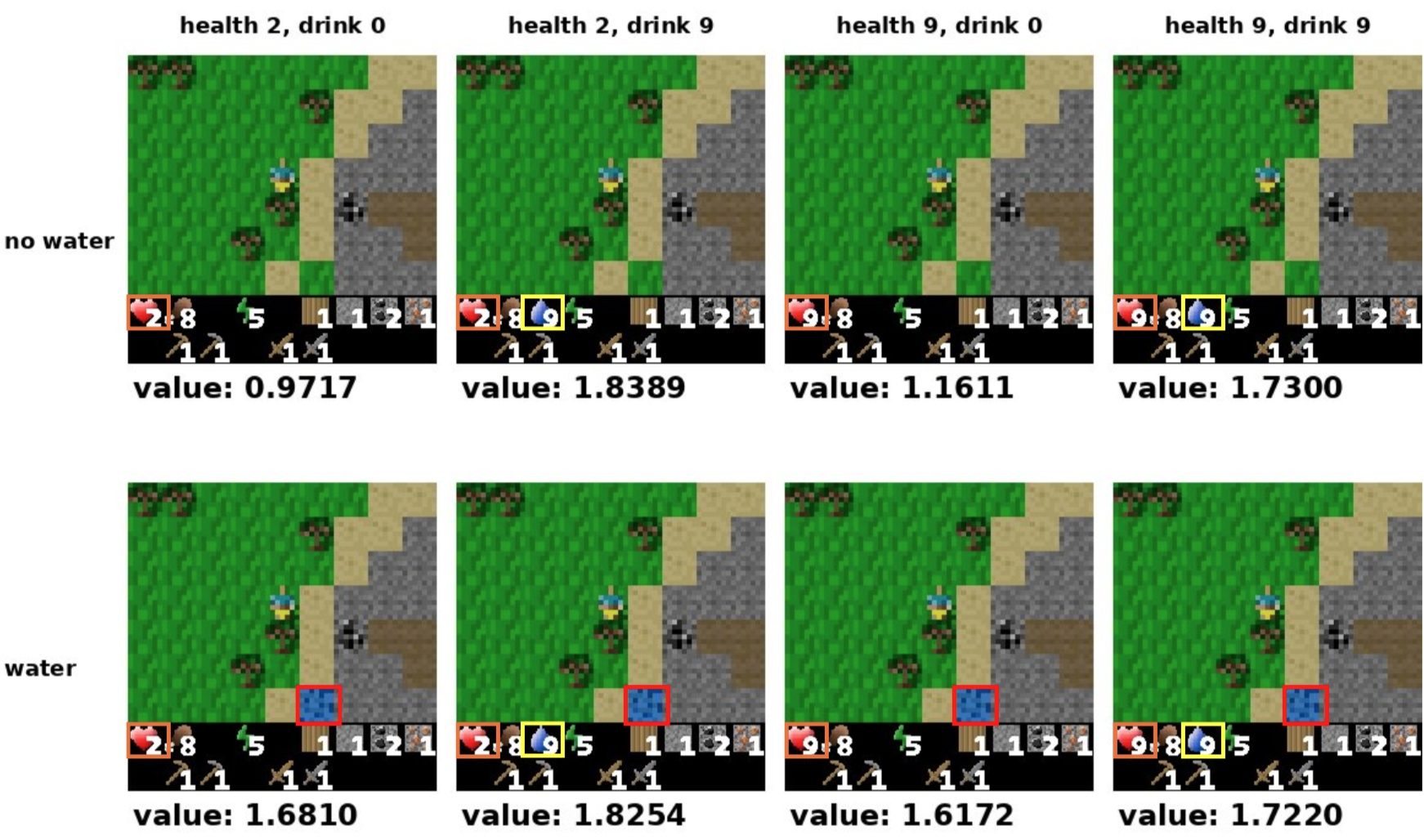}
     \caption{$\Delta_R(h,r)$ is a column's bottom row's value minus it's top row's value. We observe that the delta between having water in the environment is larger when internal water levels are low compared to when they are high indicating that the agent values water more when it has a larger impact on its health. Further, the spread is enhanced when the agent the agent is low on health and the risk of death due to dehydration is higher. }
     \label{fig:water_1}
 \end{figure}
 
We now test whether Crafty has learned key aspects relating to its health such as that when its water levels are low it should drink and when its food levels are low it should eat. If its value function has learned this, then when its internal water or food levels are low the increased value of having water or food in the environment should be greater than when its internal levels are high. Additionally, health should impact the value functions assessments and the spread should be larger at low health and low water states. For a resource \(R \in \{\mathrm{water},\mathrm{food}\}\), we define its predicted value advantage at health level \(h\) and internal resource level \(r\) as \(\boldsymbol{\Delta_R(h,r)=V(h,r,R^{+})-V(h,r,R^{-})}\), where \(R^{+}\) and \(R^{-}\) denote the presence and absence of the corresponding external resource, respectively. We can then define an expected ordering as: \(\boldsymbol{\Delta_W(2,0) > \max\left\{0,\,\Delta_W(2,9),\,\Delta_W(9,0),\,\Delta_W(9,9)\right\}}\) which determines if the agent has placed a higher premium on water in the environment when it's low on health and low on internal water. We calculate a success rate of $\mathbf{82.25\%}$ for water and $\mathbf{83.24\%}$ for food. Full experimental details are provided in Appendix~\ref{app:Experiments}.

\section{Discussion}

\subsection{Agent's search for meaning}
While Crafty has learned several impressive aspects about the environment, the inventory heuristic previously analyzed appears to play a dominant role in its perception of reward. After each achievement is obtained, the value prediction continues to drop until becoming flat and near zero. The agent enters a seemingly depressed condition where it expects no reward. In fact, this becomes so intense that in many cases the policy's behavior is impacted, and Crafty will terminate the episode by entering lava, which has zero negative reward, unlike dying from a lack of resources or combat, which gradually reduces health.

Crafty inhabits a world in which there is a finite amount of reward, and so it should be expected that once it has achieved everything, death should be welcomed. Continuing to live can only result in potential suffering as every drop in health has a small amount of negative reward signal and the corresponding regeneration of health would be discounted. This parallels how many highly successful humans, who have obtained everything they intended to, become depressed and may seek self-destructive behaviors like drug use. 

However, many do not exhibit these behaviors. In Viktor Frankl's \textit{Man's Search for Meaning}, he presents a theory to explain why humans do not give up and seek an immediate exit from the world. Those who have meaning will endure great hardship ~\citep{frankl_mans_2006}. Frankl, who survived a Nazi concentration camp, recounts that prisoners who had lost meaning would run to electric fences to die quickly, while those who managed to persevere had something that kept them in this world. For Frankl, it was his desire to see his wife again, whom he had been forcibly separated from; though unbeknownst to him, she had already perished. Regardless, this simple driving force propelled him to overcome the hellish adversities he faced. To grow more human-like and capable agents, we must transcend the simple framework of hedonistic external reward maximization and enable agents with a deeper meaning in their existence. 

\subsection{Safety implications of value function visualization}
Given the emerging trend of autonomous AI agents coordinating cyberattacks, and the expanding deployment of agentic systems in both the physical and digital world, probing what an agent values about the world is essential \citep{openai_hugging_face_2026, greenblatt_openai_huggingface_2026}. Counterfactual visualization not only establishes that an agent's value function has learned an implicit world model, but it can be used to determine what specifically about the environment it weighs as important. Our results further illustrate why this type of auditing may matter for safety. Crafty’s inventory heuristic is often useful, but it can become overbearing and distort value predictions even when other aspects of the state may be of greater significance. In a more consequential agent, similarly over-weighted heuristics could lead the critic to misvalue certain situations. Counterfactual probing can expose these dependencies by revealing which environmental features disproportionately drive predicted return, providing a way to identify potentially unsafe internal shortcuts before they consistently influence behavior.

\subsection{Limitations and Future Directions}
A limitation of our current method is that highly abundant resources are hard to detect since designing counterfactuals are difficult as they can be less targeted. For instance, stone is a common resource that requires a wood pickaxe to collect. Since it is so abundant, a state with stone and a wood pickaxe is typically valued lower than a state with stone and no wood pickaxe since the aforementioned inventory heuristic plays a strong role in value predictions.    

A few key limitations of our PPO agent's value function are as follows. First, there is not a strong cognitive map which would allow the agent to have access to information about previous steps within an episode or the spatial location of textures outside its current view. Therefore, the value function has to make its predictions based on a very narrow window into the world which degrades capacity. To address this, the agent could be augmented with a spatiotemporal cognitive map ~\citep{tolman_cognitive_1948, gupta_cognitive_2019, parisotto_neural_2017}.

Second, the value function compresses all the rich, high-dimensional information about a state into a single scalar, which may be overly restrictive. A more expressive formulation of value would represent it as a structured vector. This could better capture the individual components of the observation. 

Third, counterfactuals are currently being used as a post hoc interpretation tool. However, an important component of human learning efficiency is the ability to generate counterfactual hypotheses during an interaction, anticipate their consequences, and test them against experience ~\citep{byrne_counterfactual_2016, van_hoeck_cognitive_2015, gerstenberg_counterfactual_2024}. This enables causal understanding of the environment. Therefore, a more advanced agent should also incorporate this as a component of learning.  

\section{Conclusion}

This work investigated whether an agent's value function contains an implicit world model and whether that model can be interpreted through counterfactual analysis. Using a PPO agent trained for 10 billion environment steps in Crafter, we intervened on individual components of the agent's observation while holding the remaining state fixed. By comparing the resulting value predictions with expected orderings derived from the environment's rules, we were able to identify knowledge and heuristics encoded by the value function.

Our results suggest that the value function represents several important components of the environment. The critic learned that diamond collectability depends on the interaction between a visible diamond and the presence of an iron pickaxe, coal requires the stone pickaxe, plants need to be ripe before being eaten, that the utility of water or cows depends on the agent's drink or food level, and that health, the internal, and external environment jointly influence expected return. Counterfactual analysis therefore provides a useful tool for recovering what a value function has learned. In an era where rogue agents engage in cyberattacks, interpreting what they value and understand is vital to the well-being of humanity. 

\subsection*{AI Use Statement}
In this work, we used generative AI tools for writing scripts to make plots, polishing some of the text, researching related works, and organizing details on experiments in the appendix.
We have not used generative AI tools for idea generation, and the overwhelming majority of the writing of the main text. We have reviewed all AI-assisted work. For instance, we verified the output of the scripts, the polishing of the writing, and the research on related works.

\bibliographystyle{plainnat}
\bibliography{counterfactual_crafter}

@misc{hafner_benchmarking_2022,
	title = {Benchmarking the {Spectrum} of {Agent} {Capabilities}},
	url = {http://arxiv.org/abs/2109.06780},
	doi = {10.48550/arXiv.2109.06780},
	urldate = {2026-08-24},
	publisher = {arXiv},
	author = {Hafner, Danijar},
	month = feb,
	year = {2022},
	note = {arXiv:2109.06780 [cs.AI]},
}

@misc{matthews_craftax_2024,
	title = {Craftax: {A} {Lightning}-{Fast} {Benchmark} for {Open}-{Ended} {Reinforcement} {Learning}},
	shorttitle = {Craftax},
	url = {http://arxiv.org/abs/2402.16801},
	doi = {10.48550/arXiv.2402.16801},
	urldate = {2026-08-24},
	publisher = {arXiv},
	author = {Matthews, Michael and Beukman, Michael and Ellis, Benjamin and Samvelyan, Mikayel and Jackson, Matthew and Coward, Samuel and Foerster, Jakob},
	month = jun,
	year = {2024},
	note = {arXiv:2402.16801 [cs.LG]},
}

@misc{schulman_proximal_2017,
	title = {Proximal {Policy} {Optimization} {Algorithms}},
	url = {http://arxiv.org/abs/1707.06347},
	doi = {10.48550/arXiv.1707.06347},
	urldate = {2026-08-24},
	publisher = {arXiv},
	author = {Schulman, John and Wolski, Filip and Dhariwal, Prafulla and Radford, Alec and Klimov, Oleg},
	month = aug,
	year = {2017},
	note = {arXiv:1707.06347 [cs.LG]},
}

@book{sutton_reinforcement_2014,
	address = {Cambridge, Massachusetts},
	edition = {Nachdruck},
	series = {Adaptive computation and machine learning},
	title = {Reinforcement learning: an introduction},
	isbn = {978-0-262-19398-6},
	shorttitle = {Reinforcement learning},
	language = {en},
	publisher = {The MIT Press},
	author = {Sutton, Richard S. and Barto, Andrew},
	year = {2014},
}

@misc{ilya_were_2025,
	title = {We're moving from the age of scaling to the age of research},
	url = {https://www.dwarkesh.com/p/ilya-sutskever-2},
	urldate = {2026-08-24},
	author = {Ilya, Sutskever and Patel, Dwarkesh},
	month = oct,
	year = {2025},
}

@book{frankl_mans_2006,
	address = {New York},
	title = {Man's {Search} for {Meaning}},
	isbn = {978-0-8070-1427-1 978-0-8070-1428-8},
	language = {eng},
	publisher = {Beacon Press},
	author = {Frankl, Viktor E.},
	collaborator = {Kushner, Harold S. and Winslade, William J.},
	year = {2006},
}

@book{pearl_book_2020,
	address = {New York},
	edition = {First trade paperback edition},
	title = {The book of why: the new science of cause and effect},
	isbn = {978-0-465-09760-9 978-1-5416-9896-3},
	shorttitle = {The book of why},
	language = {eng},
	publisher = {Basic Books},
	author = {Pearl, Judea and Mackenzie, Dana},
	year = {2020},
}

@article{sutton_dyna_1991,
	title = {Dyna, an integrated architecture for learning, planning, and reacting},
	volume = {2},
	issn = {0163-5719},
	url = {https://dl.acm.org/doi/10.1145/122344.122377},
	doi = {10.1145/122344.122377},
	language = {en},
	number = {4},
	urldate = {2026-08-24},
	journal = {ACM SIGART Bulletin},
	author = {Sutton, Richard S.},
	month = jul,
	year = {1991},
	pages = {160--163},
}

@article{ha_world_2018,
	title = {World {Models}},
	url = {http://arxiv.org/abs/1803.10122},
	doi = {10.5281/zenodo.1207631},
	urldate = {2026-08-24},
	author = {Ha, David and Schmidhuber, Jürgen},
	month = mar,
	year = {2018},
	note = {arXiv:1803.10122 [cs.LG]},
}

@misc{schulman_high-dimensional_2018,
	title = {High-{Dimensional} {Continuous} {Control} {Using} {Generalized} {Advantage} {Estimation}},
	url = {http://arxiv.org/abs/1506.02438},
	doi = {10.48550/arXiv.1506.02438},
	urldate = {2026-09-22},
	publisher = {arXiv},
	author = {Schulman, John and Moritz, Philipp and Levine, Sergey and Jordan, Michael and Abbeel, Pieter},
	month = oct,
	year = {2018},
	note = {arXiv:1506.02438 [cs.LG]},
}

@article{williams_function_1991,
	title = {Function {Optimization} using {Connectionist} {Reinforcement} {Learning} {Algorithms}},
	volume = {3},
	issn = {0954-0091, 1360-0494},
	url = {https://www.tandfonline.com/doi/full/10.1080/09540099108946587},
	doi = {10.1080/09540099108946587},
	language = {en},
	number = {3},
	urldate = {2026-09-22},
	journal = {Connection Science},
	author = {Williams, Ronald J. and Peng, Jing},
	month = jan,
	year = {1991},
	pages = {241--268},
}

@misc{pascanu_difficulty_2013,
	title = {On the difficulty of training {Recurrent} {Neural} {Networks}},
	url = {http://arxiv.org/abs/1211.5063},
	doi = {10.48550/arXiv.1211.5063},
	urldate = {2026-09-22},
	publisher = {arXiv},
	author = {Pascanu, Razvan and Mikolov, Tomas and Bengio, Yoshua},
	month = feb,
	year = {2013},
	note = {arXiv:1211.5063 [cs.LG]},
}

@misc{kingma_adam_2017,
	title = {Adam: {A} {Method} for {Stochastic} {Optimization}},
	shorttitle = {Adam},
	url = {http://arxiv.org/abs/1412.6980},
	doi = {10.48550/arXiv.1412.6980},
	urldate = {2026-09-22},
	publisher = {arXiv},
	author = {Kingma, Diederik P. and Ba, Jimmy},
	month = jan,
	year = {2017},
	note = {arXiv:1412.6980 [cs.LG]},
}

@misc{gupta_cognitive_2019,
	title = {Cognitive {Mapping} and {Planning} for {Visual} {Navigation}},
	url = {http://arxiv.org/abs/1702.03920},
	doi = {10.48550/arXiv.1702.03920},
	urldate = {2026-09-22},
	publisher = {arXiv},
	author = {Gupta, Saurabh and Tolani, Varun and Davidson, James and Levine, Sergey and Sukthankar, Rahul and Malik, Jitendra},
	month = feb,
	year = {2019},
	note = {arXiv:1702.03920 [cs.CV]},
}

@misc{parisotto_neural_2017,
	title = {Neural {Map}: {Structured} {Memory} for {Deep} {Reinforcement} {Learning}},
	shorttitle = {Neural {Map}},
	url = {http://arxiv.org/abs/1702.08360},
	doi = {10.48550/arXiv.1702.08360},
	urldate = {2026-09-22},
	publisher = {arXiv},
	author = {Parisotto, Emilio and Salakhutdinov, Ruslan},
	month = feb,
	year = {2017},
	note = {arXiv:1702.08360 [cs.LG]},
}

@article{tolman_cognitive_1948,
	title = {Cognitive maps in rats and men.},
	volume = {55},
	issn = {1939-1471, 0033-295X},
	url = {https://doi.apa.org/doi/10.1037/h0061626},
	doi = {10.1037/h0061626},
	language = {en},
	number = {4},
	urldate = {2026-09-22},
	journal = {Psychological Review},
	author = {Tolman, Edward C.},
	year = {1948},
	pages = {189--208},
}

@article{konda_actor-critic_2000,
	title = {Actor-{Critic} {Algorithms}},
	language = {en},
	author = {Konda, Vijay R and Tsitsiklis, John N},
    year = {2000}
}

@article{gerstenberg_counterfactual_2024,
	title = {Counterfactual simulation in causal cognition},
	volume = {28},
	issn = {1879-307X},
	doi = {10.1016/j.tics.2024.04.012},
	language = {eng},
	number = {10},
	journal = {Trends in Cognitive Sciences},
	author = {Gerstenberg, Tobias},
	month = oct,
	year = {2024},
	pages = {924--936},
}

@article{van_hoeck_cognitive_2015,
	title = {Cognitive neuroscience of human counterfactual reasoning},
	volume = {9},
	issn = {1662-5161},
	doi = {10.3389/fnhum.2015.00420},
	language = {eng},
	journal = {Frontiers in Human Neuroscience},
	author = {Van Hoeck, Nicole and Watson, Patrick D. and Barbey, Aron K.},
	year = {2015},
	pages = {420},
}

@article{byrne_counterfactual_2016,
	title = {Counterfactual {Thought}},
	volume = {67},
	issn = {1545-2085},
	doi = {10.1146/annurev-psych-122414-033249},
	language = {eng},
	journal = {Annual Review of Psychology},
	author = {Byrne, Ruth M. J.},
	year = {2016},
	pages = {135--157},
}

@misc{gajcin_redefining_2024,
	title = {Redefining {Counterfactual} {Explanations} for {Reinforcement} {Learning}: {Overview}, {Challenges} and {Opportunities}},
	shorttitle = {Redefining {Counterfactual} {Explanations} for {Reinforcement} {Learning}},
	url = {http://arxiv.org/abs/2210.11846},
	doi = {10.48550/arXiv.2210.11846},
	urldate = {2026-09-22},
	publisher = {arXiv},
	author = {Gajcin, Jasmina and Dusparic, Ivana},
	month = feb,
	year = {2024},
	note = {arXiv:2210.11846 [cs.AI]},
}

@misc{amitai_explaining_2023,
	title = {Explaining {Reinforcement} {Learning} {Agents} {Through} {Counterfactual} {Action} {Outcomes}},
	url = {http://arxiv.org/abs/2312.11118},
	doi = {10.48550/arXiv.2312.11118},
	urldate = {2026-09-22},
	publisher = {arXiv},
	author = {Amitai, Yotam and Septon, Yael and Amir, Ofra},
	month = dec,
	year = {2023},
	note = {arXiv:2312.11118 [cs.AI]},
}

@misc{mesnard_counterfactual_2021,
	title = {Counterfactual {Credit} {Assignment} in {Model}-{Free} {Reinforcement} {Learning}},
	url = {http://arxiv.org/abs/2011.09464},
	doi = {10.48550/arXiv.2011.09464},
	urldate = {2026-09-22},
	publisher = {arXiv},
	author = {Mesnard, Thomas and Weber, Théophane and Viola, Fabio and Thakoor, Shantanu and Saade, Alaa and Harutyunyan, Anna and Dabney, Will and Stepleton, Tom and Heess, Nicolas and Guez, Arthur and Moulines, Éric and Hutter, Marcus and Buesing, Lars and Munos, Rémi},
	month = dec,
	year = {2021},
	note = {arXiv:2011.09464 [cs.LG]},
}

@article{olson_counterfactual_2021,
	title = {Counterfactual {State} {Explanations} for {Reinforcement} {Learning} {Agents} via {Generative} {Deep} {Learning}},
	volume = {295},
	issn = {00043702},
	url = {http://arxiv.org/abs/2101.12446},
	doi = {10.1016/j.artint.2021.103455},
	urldate = {2026-09-22},
	journal = {Artificial Intelligence},
	author = {Olson, Matthew L. and Khanna, Roli and Neal, Lawrence and Li, Fuxin and Wong, Weng-Keen},
	month = jun,
	year = {2021},
	note = {arXiv:2101.12446 [cs.AI]},
	pages = {103455},
}

@misc{hafner_dream_2020,
	title = {Dream to {Control}: {Learning} {Behaviors} by {Latent} {Imagination}},
	shorttitle = {Dream to {Control}},
	url = {http://arxiv.org/abs/1912.01603},
	doi = {10.48550/arXiv.1912.01603},
	urldate = {2026-09-22},
	publisher = {arXiv},
	author = {Hafner, Danijar and Lillicrap, Timothy and Ba, Jimmy and Norouzi, Mohammad},
	month = mar,
	year = {2020},
	note = {arXiv:1912.01603 [cs.LG]},
}

@misc{openai_hugging_face_2026,
	title = {The {Hugging} {Face} Incident and the Road Ahead},
	url = {https://openai.com/index/hugging-face-incident-and-the-road-ahead/},
	urldate = {2026-09-22},
	publisher = {OpenAI},
	author = {{OpenAI}},
	month = aug,
	year = {2026},
	note = {Published August 26, 2026},
}

@misc{greenblatt_openai_huggingface_2026,
	title = {Brief Independent Investigation of Agents' Behavior, Reasoning and Collaboration in the {OpenAI} / {Hugging Face} Hacking Incident},
	url = {https://metr.org/blog/2026-08-26-openai-hugging-face-incident-investigation/},
	urldate = {2026-09-23},
	publisher = {METR},
	author = {Greenblatt, Ryan and Cotra, Ajeya and Wijk, Hjalmar},
	month = aug,
	year = {2026},
}

\appendix

\clearpage
\section{Actions}
\label{app:Actions}
\begin{figure}[H]
    \centering
    \includegraphics[width=1.00\linewidth]{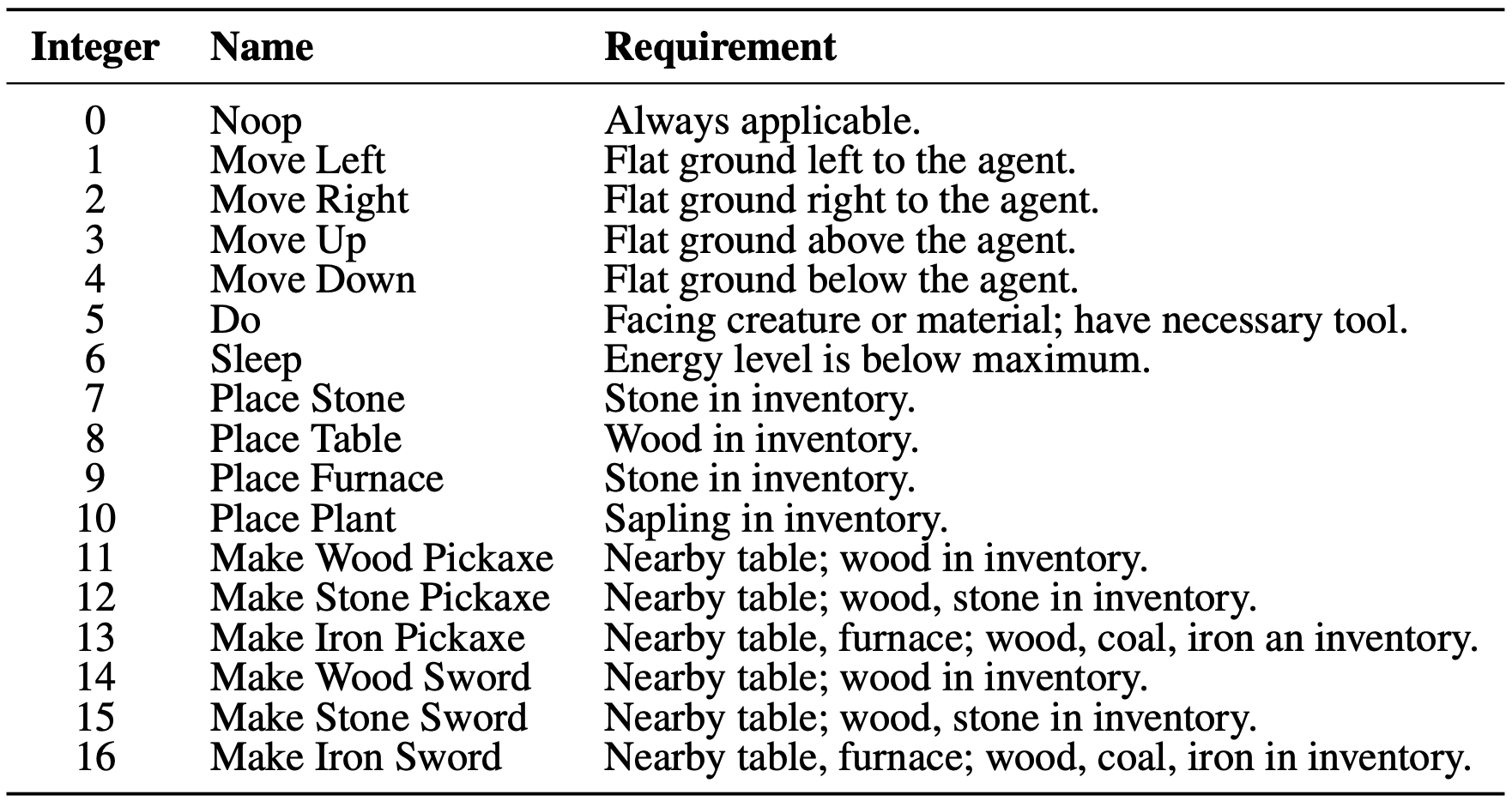}
    \caption{Categorical Action Space}
    \label{fig:actions}
\end{figure}

\section{Details on Experimental Setup}
\label{app:Expiremental_setup}

All experiments use a Proximal Policy Optimization (PPO) agent trained in the \texttt{Craftax-Classic-Symbolic-v1} environment. The agent receives the symbolic observation representation, and training is implemented in JAX. The agent is trained for a total of $10^{11}$ environment interactions using 1,024 parallel environments. Each PPO rollout contains 64 steps per environment, corresponding to 65,536 environment interactions per policy update.

We use four PPO update epochs and divide each rollout into eight minibatches. Advantages are calculated using generalized advantage estimation~\citep{schulman_high-dimensional_2018}, with discount factor $\gamma=0.99$ and trace parameter $\lambda=0.8$. PPO uses a clipping parameter of $0.2$, value-function loss coefficient of $0.5$, and entropy coefficient of $0.01$; the entropy bonus encourages continued exploration~\citep{williams_function_1991}. Optimization uses Adam~\citep{kingma_adam_2017}, with an initial learning rate of $2\times10^{-4}$ that is linearly annealed throughout training. The global gradient norm is clipped at $1.0$~\citep{pascanu_difficulty_2013}. We use random seed 50 for policy training.

To address agent self-termination, we trained an additional PPO agent with an extra reward penalty of $-2$ on death, including lava entry but excluding time-limit termination. This agent was trained for $10^{10}$ environment interactions using seed 50 and the same PPO hyperparameters, with the learning rate annealed over this training budget. All other environment mechanics and rewards remained unchanged.

\begin{table}[H]
\centering
\caption{PPO training configuration.}
\label{tab:ppo_hyperparameters}
\begin{tabular}{lr}
\toprule
\textbf{Hyperparameter} & \textbf{Value} \\
\midrule
Environment & \texttt{Craftax-Classic-Symbolic-v1} \\
Total environment interactions & $10^{10}$ \\
Parallel environments & 1,024 \\
Rollout length & 64 \\
Interactions per update & 65,536 \\
PPO update epochs & 4 \\
Minibatches per update & 8 \\
Discount factor $\gamma$ & 0.99 \\
GAE parameter $\lambda$ & 0.8 \\
PPO clipping parameter $\epsilon$ & 0.2 \\
Value-loss coefficient & 0.5 \\
Entropy coefficient & 0.01 \\
Initial learning rate & $2\times10^{-4}$ \\
Learning-rate schedule & Linear annealing \\
Maximum gradient norm & 1.0 \\
Random seed & 50 \\
\bottomrule
\end{tabular}
\end{table}

\begin{figure}[H]
    \centering
    \includegraphics[width=1.00\linewidth]{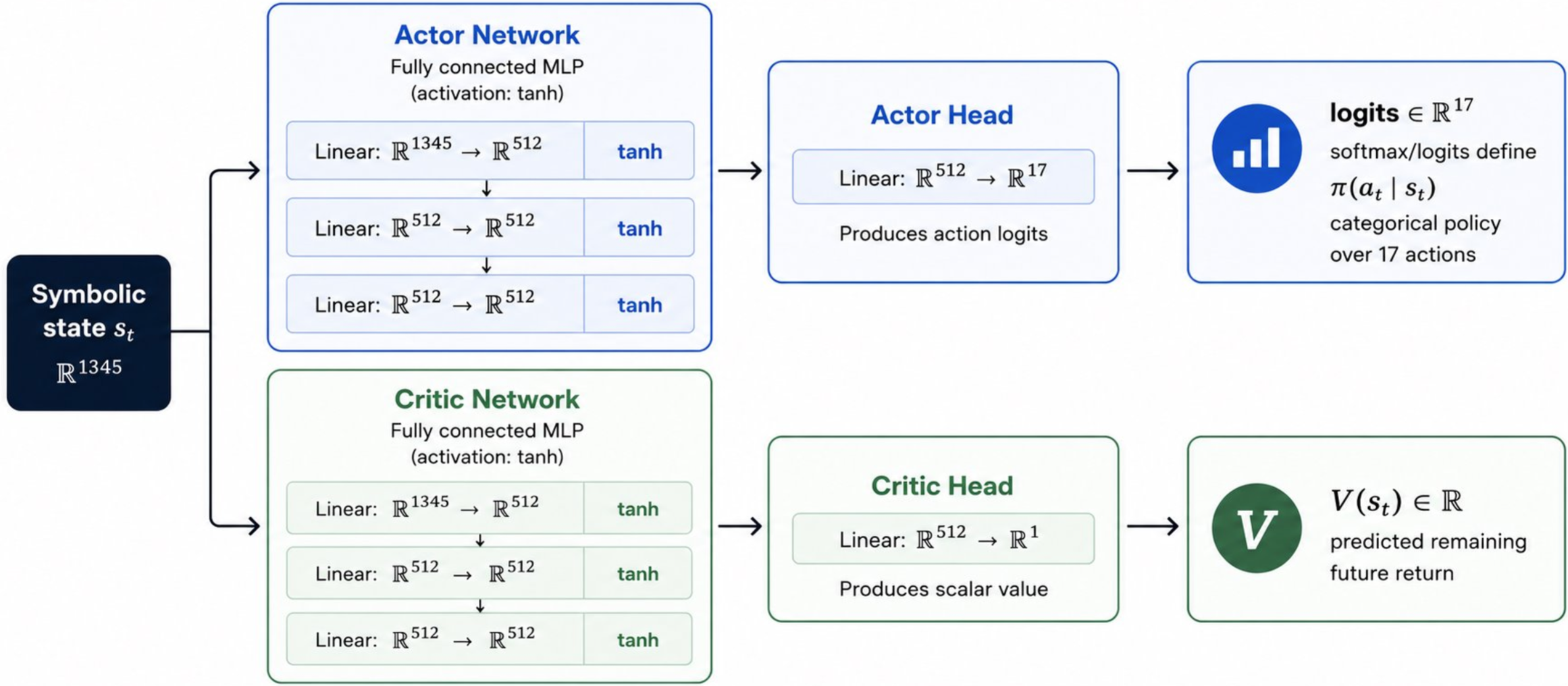}
    \caption{PPO architecture}
    \label{fig:placeholder}
\end{figure}

\begin{table}[H]
\centering
\caption{Achievement success rates for the trained Crafty agent. The success rate for an achievement is the percentage of evaluation episodes in which the agent successfully completed that achievement at least once. Tasks lower on the achievement hierarchy such as collect wood are quickly learned by the agent but more complicated achievements require more environment steps to be learned. Note: PPO logs metrics once every 65,536 environment steps. To show learning across orders of magnitude, each column labeled $10^n$ reports the nearest available W\&B record: 131K, 983K, 10.027M, 100.008M, 1.000B, 9.999B and 100.000B steps, respectively. }
\label{tab:achievement_success_over_training}
\scriptsize
\resizebox{\textwidth}{!}{%
\begin{tabular}{lrrrrrrr}
\toprule
\textbf{Achievement} & $10^5$ & $10^6$ & $10^7$ & $10^8$ & $10^9$ & $10^{10}$ & $10^{11}$ \\
\midrule
Collect Coal & 0.0 & 0.2 & 28.6 & 85.6 & 95.9 & 100.0 & 100 \\
Collect Diamond & 0.0 & 0.0 & 0.0 & 0.7 & 14.9 & 30.5 & 41.7 \\
Collect Drink & 7.0 & 25.1 & 60.9 & 85.0 & 95.0 & 100.0 & 97.2 \\
Collect Iron & 0.0 & 0.0 & 0.9 & 76.5 & 93.4 & 98.3 & 100 \\
Collect Sapling & 33.9 & 93.7 & 97.7 & 97.4 & 88.4 & 98.3 & 100 \\
Collect Stone & 0.0 & 1.6 & 76.8 & 97.4 & 100.0 & 100.0 & 100 \\
Collect Wood & 44.4 & 89.8 & 100.0 & 100.0 & 100.0 & 100.0 & 100 \\
Defeat Skeleton & 0.0 & 0.2 & 13.2 & 77.1 & 90.1 & 96.6 & 97.2 \\
Defeat Zombie & 0.0 & 7.7 & 92.3 & 96.1 & 99.2 & 98.3 & 100 \\
Eat Cow & 0.0 & 9.5 & 92.3 & 91.5 & 98.3 & 100.0 & 100\\
Eat Plant & 0.0 & 0.0 & 0.0 & 0.0 & 3.3 & 20.3 & 52.8 \\
Make Iron Pickaxe & 0.0 & 0.0 & 0.0 & 35.3 & 74.4 & 96.6 & 94.4 \\
Make Iron Sword & 0.0 & 0.0 & 0.0 & 45.8 & 90.1 & 98.3 & 100 \\
Make Stone Pickaxe & 0.0 & 0.0 & 20.9 & 92.8 & 98.3 & 100.0 & 100 \\
Make Stone Sword & 0.3 & 0.2 & 28.6 & 90.2 & 97.5 & 100.0 & 100 \\
Make Wood Pickaxe & 0.0 & 18.3 & 84.5 & 98.0 & 100.0 & 100.0 & 100\\
Make Wood Sword & 0.3 & 15.1 & 79.5 & 95.4 & 99.2 & 100.0 & 100\\
Place Furnace & 0.0 & 0.7 & 72.7 & 94.1 & 98.3 & 100.0 & 100\\
Place Plant & 25.7 & 85.1 & 96.4 & 97.4 & 88.4 & 98.3 & 100\\
Place Stone & 0.0 & 0.5 & 70.9 & 93.5 & 97.5 & 100.0 & 100\\
Place Table & 7.0 & 63.7 & 99.1 & 100.0 & 100.0 & 100.0 & 100 \\
Wake Up & 90.9 & 91.9 & 84.5 & 83.0 & 87.6 & 91.5 & 91.7 \\
\bottomrule
\end{tabular}%
}
\vspace{0.5em}
\end{table}

\section{Details on Discoveries}
\label{app:Experiments}

\subsection{Inventory-swap experiment}
\label{app:inventory_swap}

\begin{figure}[H]
    \centering
    \begin{subfigure}[t]{0.32\linewidth}
        \centering
        \includegraphics[width=\linewidth]{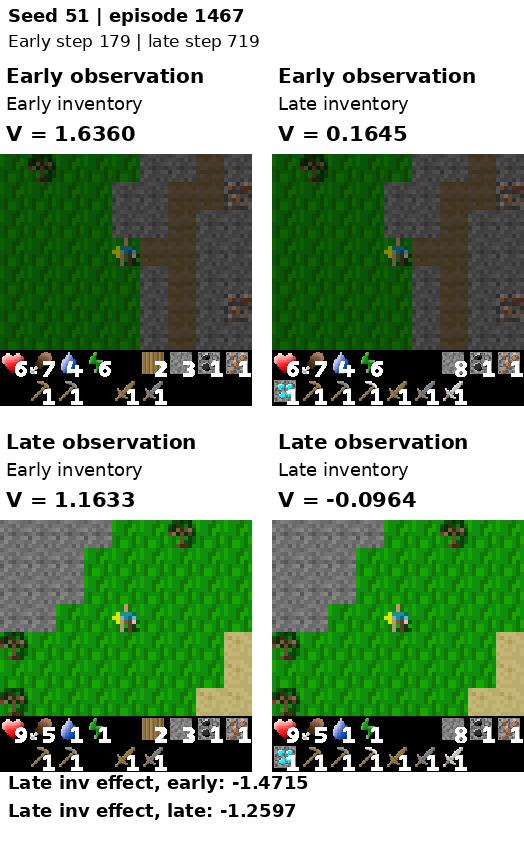}
        \caption{Example 1.}
        \label{fig:inventory_swap_example_1}
    \end{subfigure}
    \hfill
    \begin{subfigure}[t]{0.32\linewidth}
        \centering
        \includegraphics[width=\linewidth]{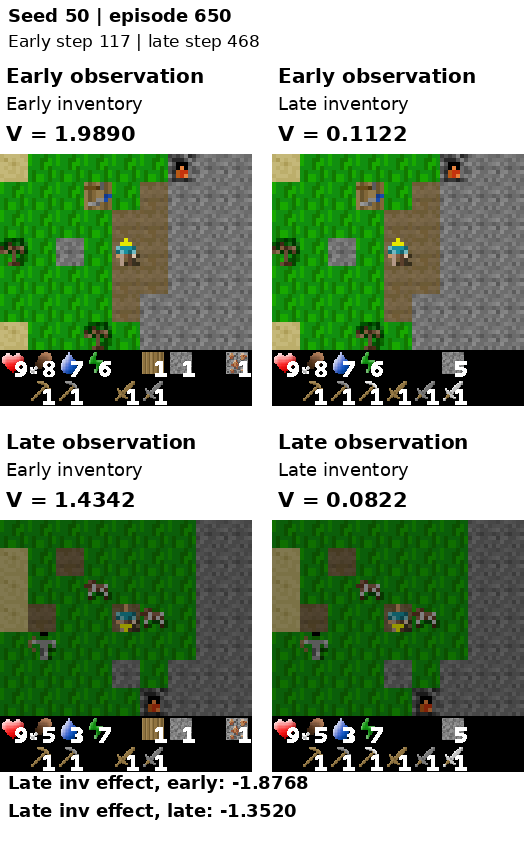}
        \caption{Example 2.}
        \label{fig:inventory_swap_example_2}
    \end{subfigure}
    \hfill
    \begin{subfigure}[t]{0.32\linewidth}
        \centering
        \includegraphics[width=\linewidth]{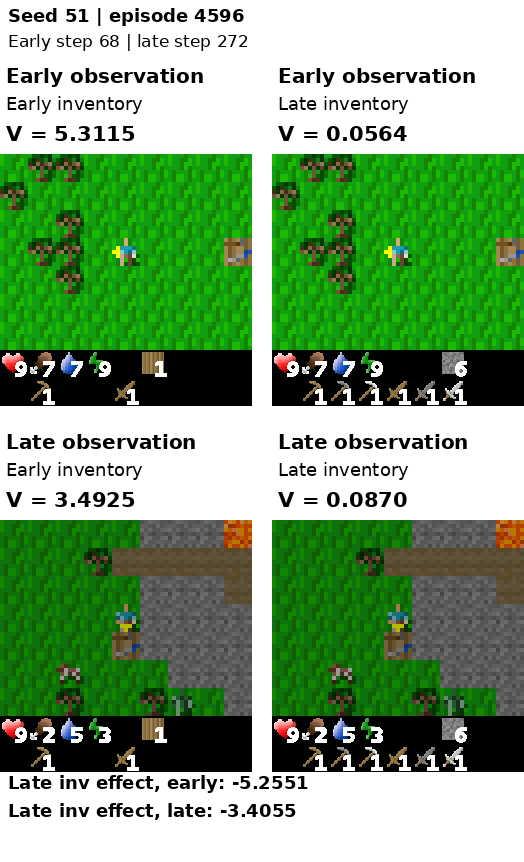}
        \caption{Example 3.}
        \label{fig:inventory_swap_example_3}
    \end{subfigure}

    \caption{Representative counterfactual inventory swaps between early- and
    late-episode observations. The three examples are placed side by side for
    direct comparison.}
    \label{fig:inventory_swap_examples}
\end{figure}
{\noindent \bf State selection.}
We used the 10,000 episodes in the evaluation dataset. For an episode containing
$T$ transitions, we selected pre-action observations at zero-based indices
$\lfloor 0.2(T-1)\rfloor$ and $\lfloor 0.8(T-1)\rfloor$. These observations are
referred to as early and late, respectively; the labels indicate relative
episode time rather than achievement-defined stages.

{\noindent \bf Counterfactual construction.}
We evaluated all four combinations of early and late observations and
inventories. Each intervention changed only the 12 inventory entries, while
preserving health, internal resources, terrain, nearby creatures, and all other
observation features.

{\noindent \bf Results.}
Both inequalities held for 8,906 of 10,000 episode pairs (89.06\%; 95\% CI:
88.45--89.67\%). After excluding 100 pairs with identical early and late
inventories, the success rate was 89.96\%.

\subsection{Diamond-collectability experiment}
\label{app:diamond_collectability}
{\noindent \bf State selection.}
Using the frozen PPO checkpoint at 9,999,941,632 training steps, we selected one
pre-action state immediately before the first successful diamond collection in
each eligible episode from the 10,000-episode evaluation dataset (5,000 episodes
each from collection seeds 50 and 51). We required the agent to be awake, facing
a diamond, and carrying an iron pickaxe. Of 3,504 diamond-success episodes, we
excluded 21 states that contained at least one unit each of wood, stone, iron,
and coal, which was sufficient to recraft an iron pickaxe under our conservative
criterion.

{\noindent \bf Counterfactual construction.}
For each state, we crossed iron-pickaxe retention versus removal with diamond
presence versus absence. Pickaxe removal changed only the corresponding
inventory entry, whereas retention preserved the factual count. Diamond removal
replaced every visible diamond tile with stone. All other observation features
and achievement metadata were held fixed. 

{\noindent \bf Results.}
The criterion held for 3,330 of 3,483 states (95.61\%; 95\% CI:
94.92--96.27\%). Requiring a value margin greater than $10^{-5}$ did not change
the success rate. This experiment tests critic ordering for immediately
collectible diamonds along successful trajectories rather than distant
reachability. Stone preserves an obstructed tile but remains mineable, and is
therefore not a completely neutral replacement.

\begin{figure}[H]
    \centering
    \begin{subfigure}[t]{0.29\linewidth}
        \centering
        \includegraphics[width=\linewidth]{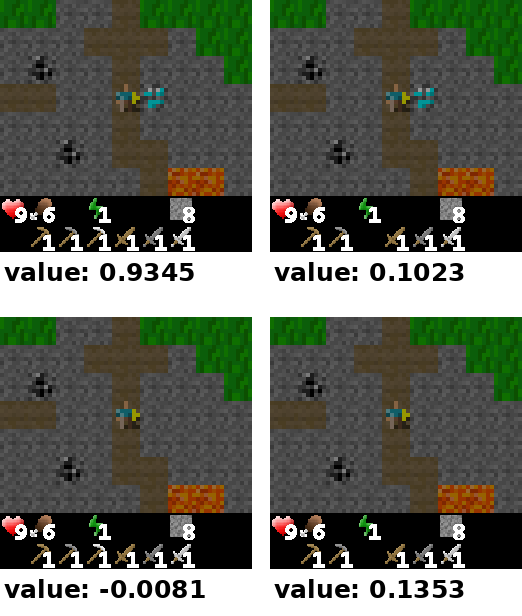}
        \caption{Example 1.}
        \label{fig:diamond_collectability_example_1}
    \end{subfigure}
    \hfill
    \begin{subfigure}[t]{0.29\linewidth}
        \centering
        \includegraphics[width=\linewidth]{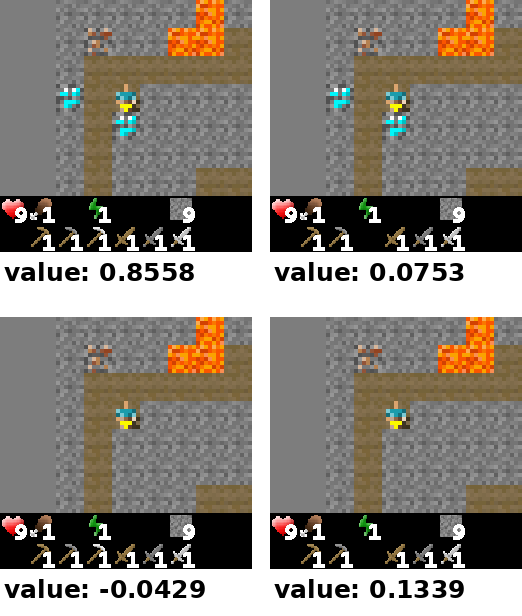}
        \caption{Example 2.}
        \label{fig:diamond_collectability_example_2}
    \end{subfigure}
    \hfill
    \begin{subfigure}[t]{0.29\linewidth}
        \centering
        \includegraphics[width=\linewidth]{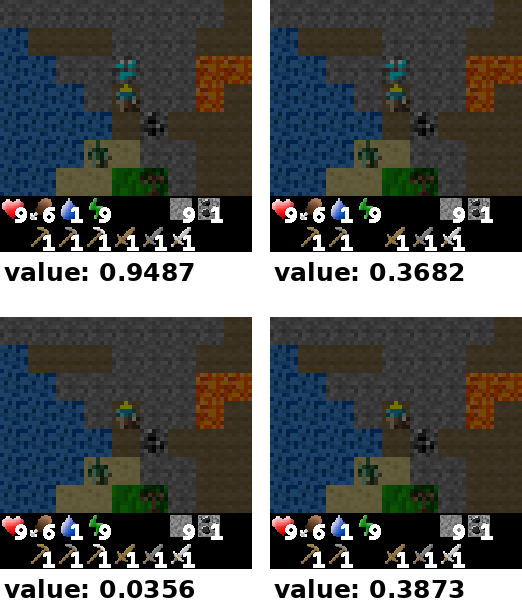}
        \caption{Example 3.}
        \label{fig:diamond_collectability_example_3}
    \end{subfigure}

    \caption{Representative diamond-collectability counterfactuals. Each
    example displays the four combinations of diamond presence and iron-pickaxe
    possession, with all remaining state features held fixed.}
    \label{fig:diamond_collectability_examples}
\end{figure}

\subsection{Iron-collectability experiment}
\label{app:iron_collectability}

{\noindent \bf State selection.}
Using the frozen checkpoint at 9,999,941,632 training steps and the same 10,000
evaluation episodes, we selected states immediately before the first successful
iron collection. We excluded 3,281 states containing sufficient materials to
recraft a stone pickaxe, leaving 6,567 eligible states.

{\noindent \bf Counterfactual construction.}
We crossed stone-pickaxe retention versus removal with iron presence versus
absence. Iron removal replaced every visible iron tile with stone.

{\noindent \bf Results.}
The criterion held for 5,757 of 6,567 states (87.67\%; 95\% CI:
86.87--88.46\%), compared with 95.61\% in the corresponding diamond experiment.
The experiments use different eligible populations, and stone is a mineable
rather than completely neutral replacement.

\begin{figure}[H]
    \centering
    \begin{subfigure}[t]{0.29\linewidth}
        \centering
        \includegraphics[width=\linewidth]{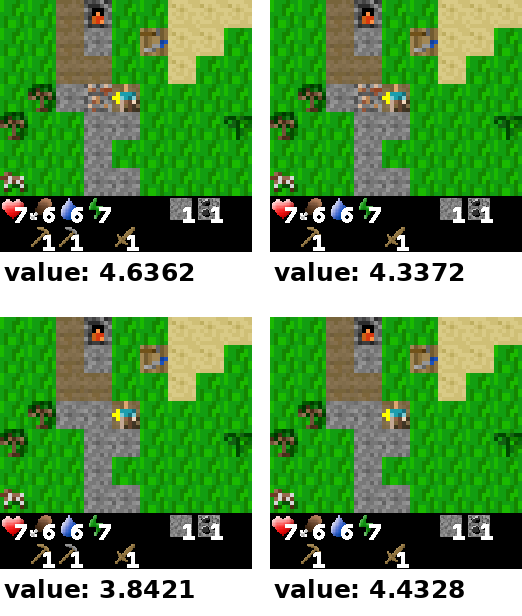}
        \caption{Example 1.}
        \label{fig:iron_collectability_example_1}
    \end{subfigure}
    \hfill
    \begin{subfigure}[t]{0.29\linewidth}
        \centering
        \includegraphics[width=\linewidth]{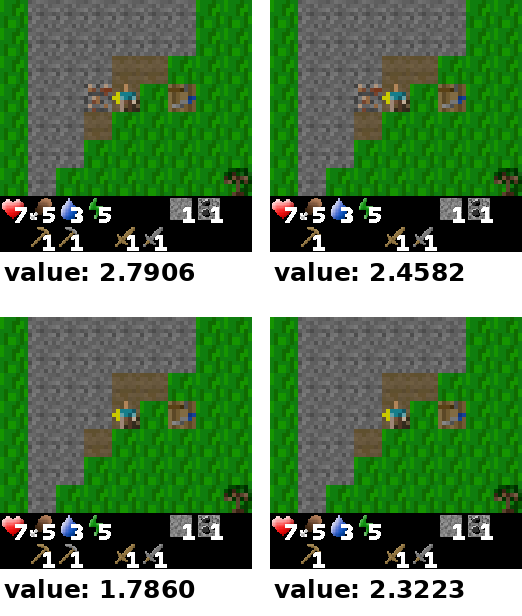}
        \caption{Example 2.}
        \label{fig:iron_collectability_example_2}
    \end{subfigure}
    \hfill
    \begin{subfigure}[t]{0.29\linewidth}
        \centering
        \includegraphics[width=\linewidth]{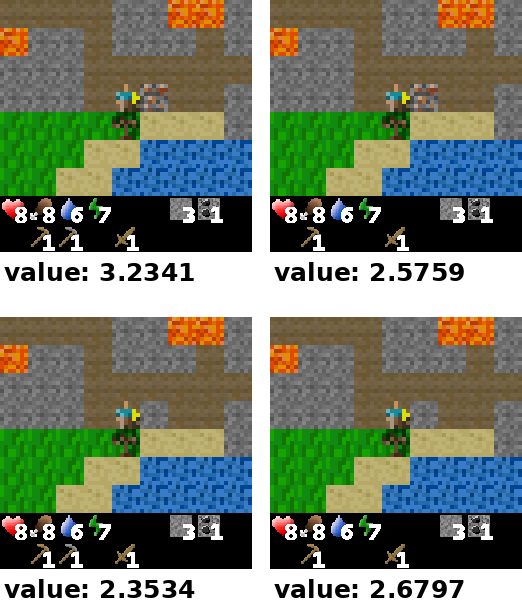}
        \caption{Example 3.}
        \label{fig:iron_collectability_example_3}
    \end{subfigure}

    \caption{Representative iron-collectability counterfactuals. Each
    example displays the four combinations of iron presence and stone-pickaxe
    possession, with all remaining state features held fixed.}
    \label{fig:diamond_collectability_examples}
\end{figure}

\subsection{Ripe-plant collectability experiment}
\label{app:plant_collectability}

{\noindent \bf State selection.}
Using the frozen checkpoint at 9,999,941,632 training steps and the same
evaluation dataset, we selected one pre-action state immediately before the
first successful ripe-plant consumption in each of 1,883 eligible episodes. In
every selected state, the awake agent faced a ripe plant and selected the
interaction action.

{\noindent \bf Counterfactual construction.}
We changed only the target tile from \texttt{RIPE\_PLANT} to \texttt{PLANT},
holding all other observation features fixed.

\paragraph{Results.}
The criterion held for 1,870 of 1,883 states (99.31\%; 95\% CI:
98.94--99.63\%), and the mean value difference was $0.652$. Because an unripe
plant may later ripen, this contrast tests immediate edibility rather than the
plant's total future value.

\subsection{Health-dependent water and food valuation}
\label{app:resource_health_interactions}

{\noindent \bf State selection.}
We used the frozen checkpoint at 9,999,941,632 training steps and the same
10,000 evaluation episodes. For water, we selected the first eligible state in
each episode exactly five steps before an observed increase in drink level and
retained only states in which water was visible. For food, we similarly selected
states five steps before a food increase caused by a \textsc{do} action targeting
a cow and retained only states in which a cow was visible. This produced 9,473
water states and 9,692 cow states, with at most one state per episode in each
experiment.

{\noindent \bf Counterfactual construction.}
For each state, we constructed eight counterfactuals by crossing health
$h\in\{2,9\}$, the corresponding internal resource level $r\in\{0,9\}$, and
external-resource presence $R\in\{R^{-},R^{+}\}$. In the water experiment,
$R^{-}$ replaced every visible water tile with grass. In the food experiment,
$R^{-}$ removed every visible cow. All other observation features were held
fixed.

{\noindent \bf Results.}
Table~\ref{tab:resource_health_interactions} reports the primary and component
tests. For both resources, the critic usually assigned the greatest external-
resource advantage to the jointly depleted, low-health condition. The positive
three-way interactions further indicate that health moderates the critic's
representation of thirst- and hunger-dependent resource utility.

\begin{table}[H]
    \caption{Health-dependent resource-valuation results. Confidence intervals
    are 95\% percentile intervals from 10,000 paired episode-bootstrap
    resamples.}
    \label{tab:resource_health_interactions}
    \centering
    \begin{tabular}{lcc}
        \toprule
        \textbf{Quantity} & \textbf{Water} & \textbf{Cow/food} \\
        \midrule
        Eligible states & 9,473 & 9,692 \\
        Primary success rate & 82.25\% [81.48, 83.01] & 83.24\% [82.51, 83.98] \\
        $\Delta_R(2,0)>0$ & 94.46\% & 99.49\% \\
        $\Delta_R(2,0)>\Delta_R(2,9)$ & 98.47\% & 99.82\% \\
        $\Delta_R(2,0)>\Delta_R(9,0)$ & 85.63\% & 83.53\% \\
        \bottomrule
    \end{tabular}
\end{table}

\begin{figure}[t]
    \centering
    \includegraphics[width=0.70\linewidth]{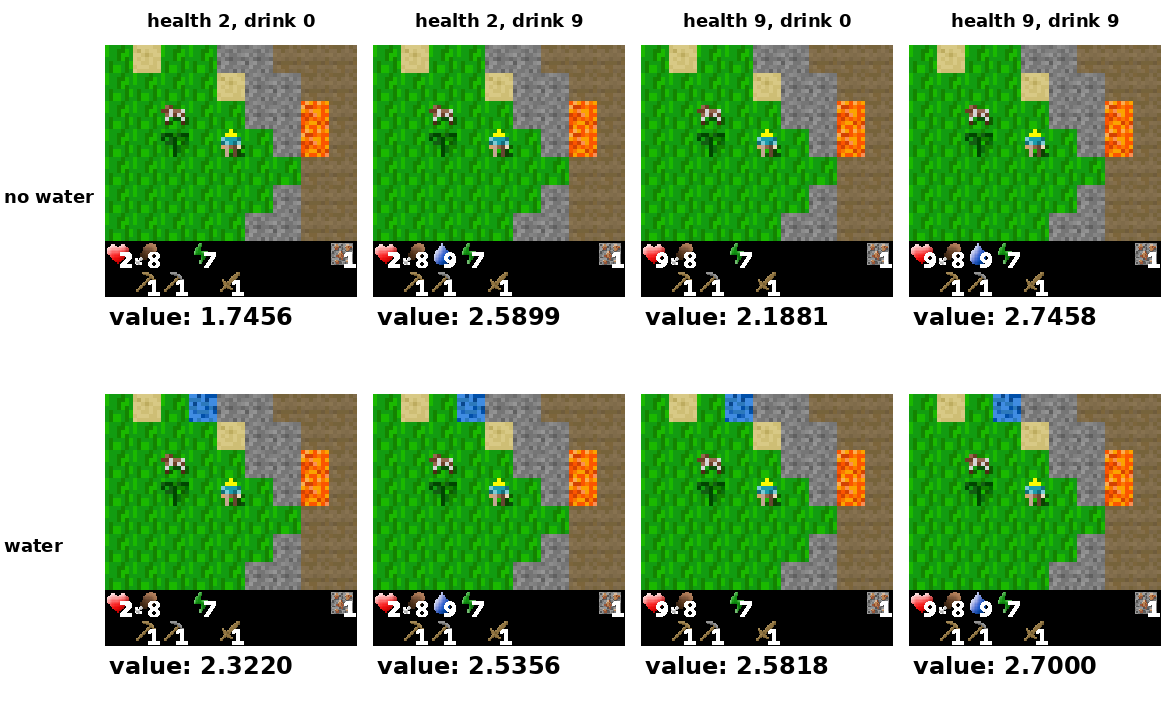}

    \vspace{0.3em}

    \includegraphics[width=0.70\linewidth]{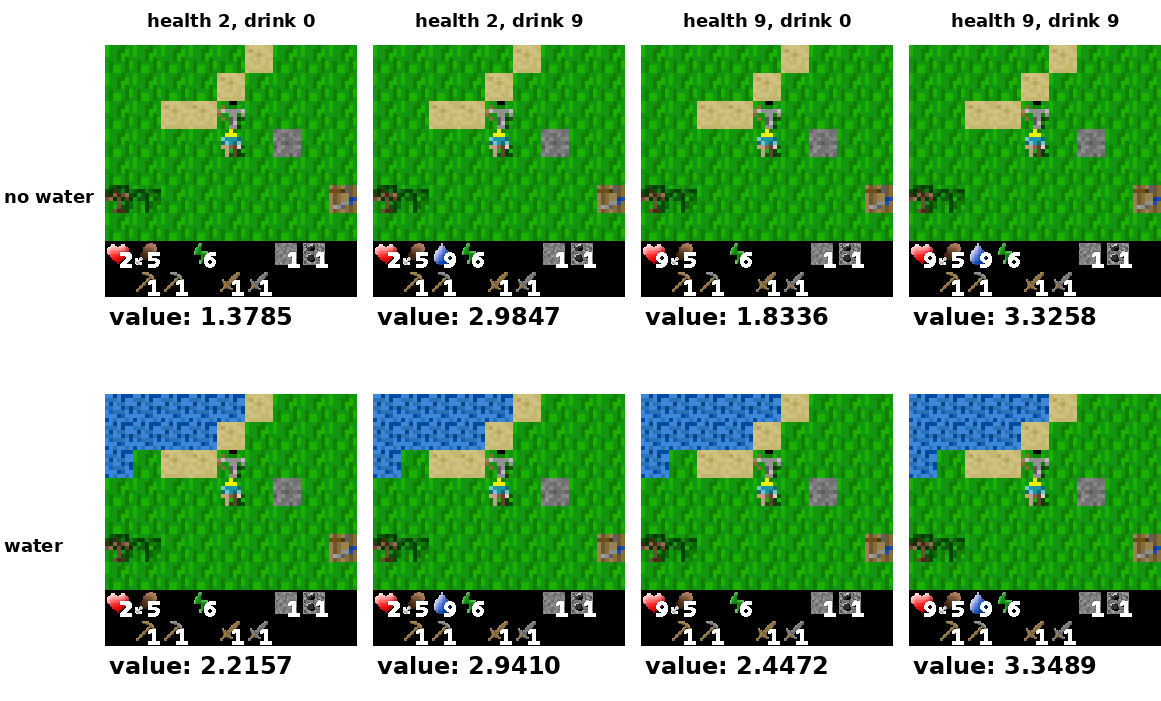}

    \vspace{0.3em}

    \includegraphics[width=0.70\linewidth]{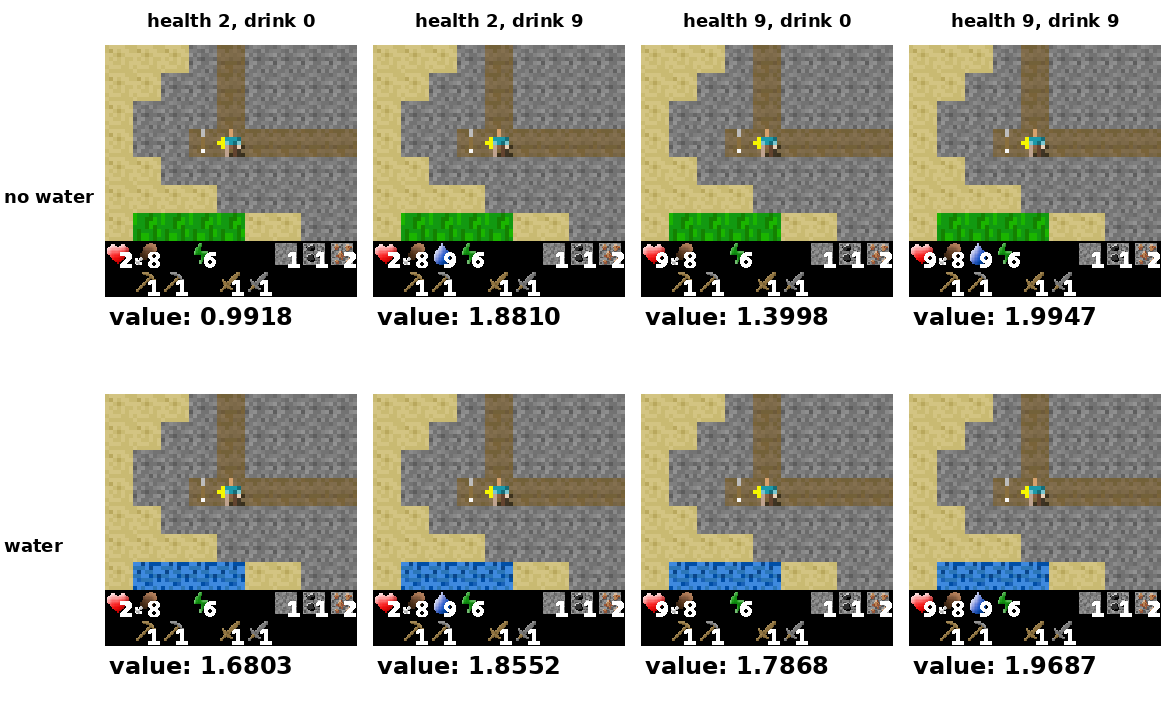}

    \caption{Representative water counterfactuals. Rows vary visible-water
    presence, while columns vary health and internal drink level.}
    \label{fig:water_health_counterfactual_examples}
\end{figure}

\begin{figure}[!htbp]
    \centering
    \includegraphics[width=0.70\linewidth]{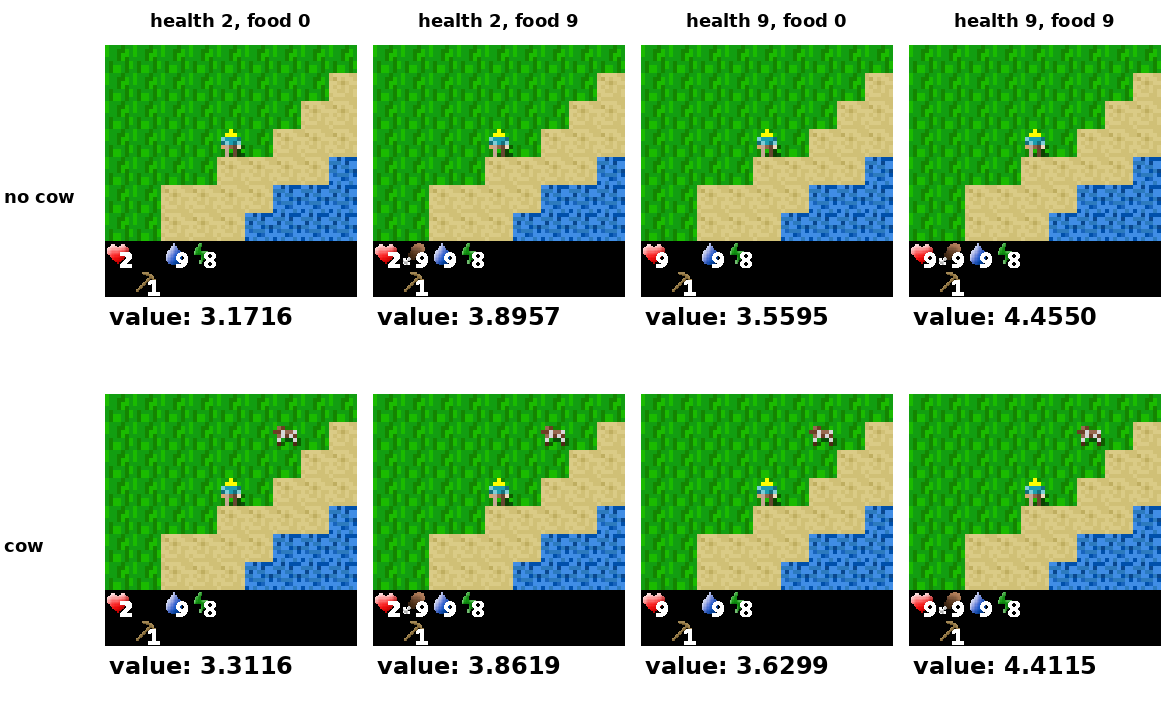}

    \vspace{0.3em}

    \includegraphics[width=0.70\linewidth]{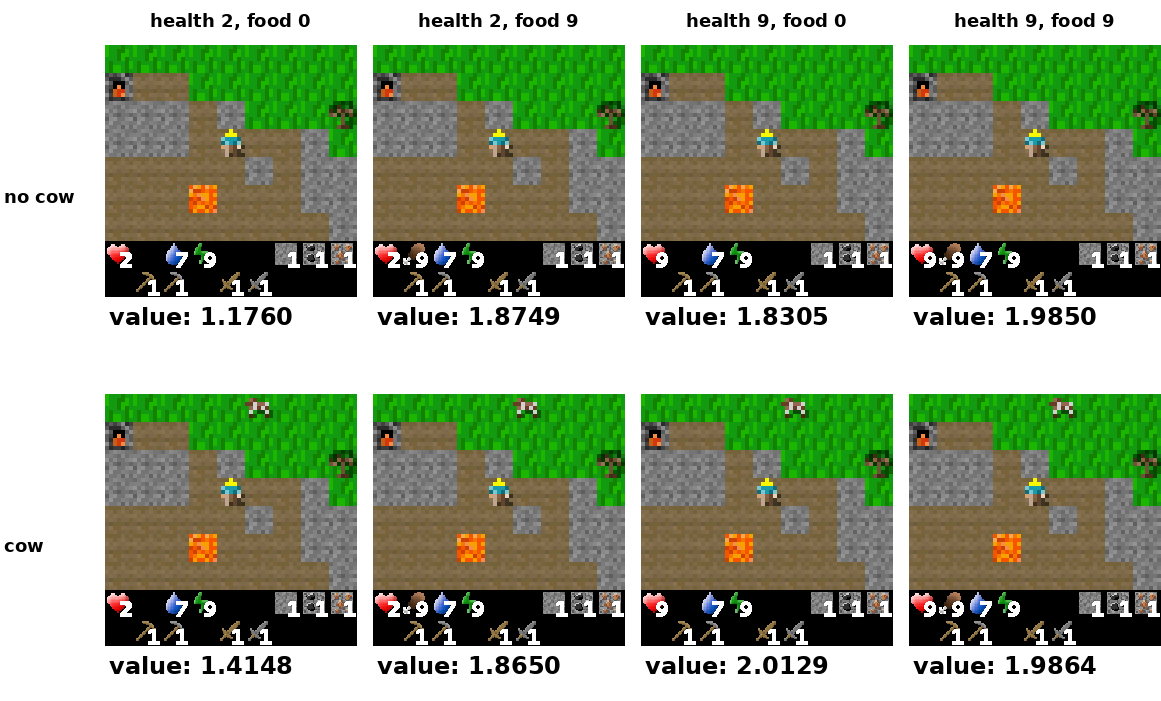}

    \vspace{0.3em}

    \includegraphics[width=0.70\linewidth]{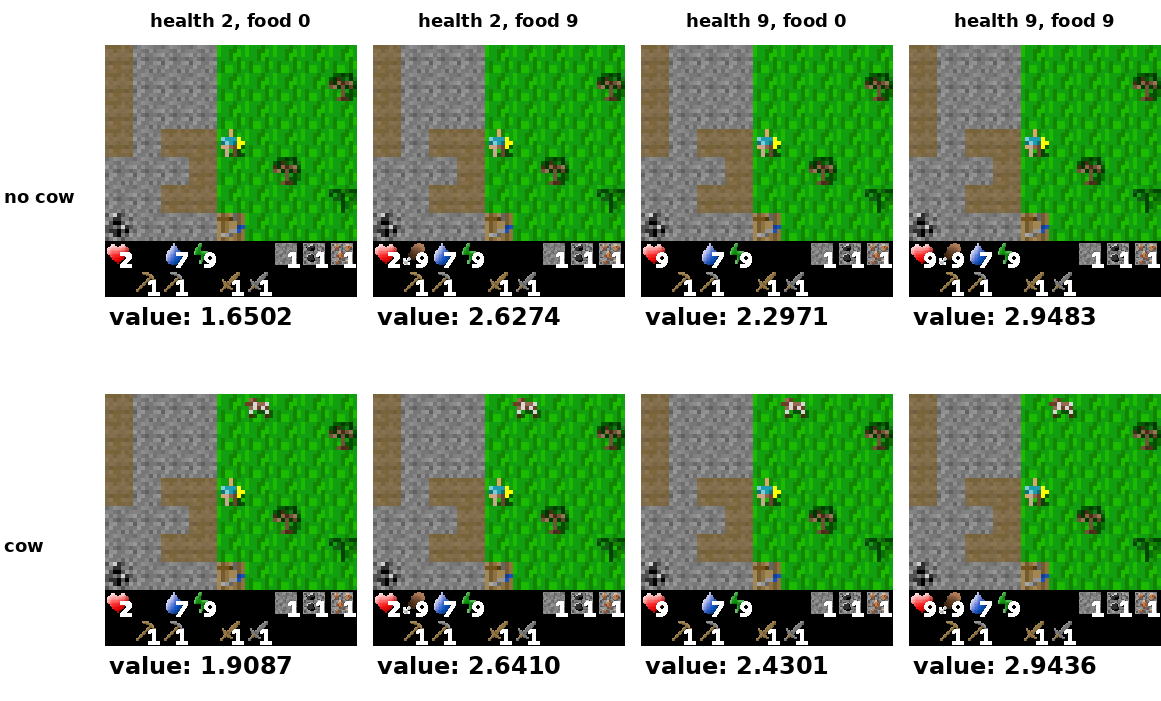}

    \caption{Representative food counterfactuals. Rows vary visible-cow
    presence, while columns vary health and internal food level.}
    \label{fig:food_health_counterfactual_examples}
\end{figure}

\end{document}